\documentclass[11pt]{article}
\usepackage[final]{acl}
\usepackage{times}
\usepackage{latexsym}
\usepackage[T1]{fontenc}
\usepackage[utf8]{inputenc}
\usepackage{microtype}
\usepackage{inconsolata}
\usepackage{graphicx}
\usepackage{booktabs}
\usepackage{amsmath}
\usepackage{amssymb}
\usepackage[table]{xcolor}
\usepackage{subfig}
\usepackage{multicol}
\usepackage{tcolorbox}
\usepackage{listings}
\usepackage{makecell}
\usepackage[normalem]{ulem}

\tcbuselibrary{skins,breakable}
\usepackage{xcolor}
\usepackage{array}

\definecolor{errorbg}{RGB}{255,232,232}
\definecolor{tracegray}{RGB}{248,248,248}

\definecolor{groupA}{HTML}{E8F3FF} 
\definecolor{groupB}{HTML}{F7F7F7} 
\definecolor{groupC}{HTML}{EAF9EB}
\definecolor{groupD}{HTML}{FFF7E6}

\title{AdaC-GRPO: Towards Mastering Long-Horizon Visual Sudoku Puzzles via Adaptive Curriculum Learning}

\author{
\textbf{Sida Li}\textsuperscript{1} \quad
\textbf{Yuanxin Liu}\textsuperscript{1} \quad
\textbf{Linli Yao}\textsuperscript{1} \quad
\textbf{Shicheng Li}\textsuperscript{1} \quad
\textbf{Xu Sun}\textsuperscript{1} \\
\textbf{Haoli Bai}\textsuperscript{2} \quad
\textbf{Xiao-Hui Li}\textsuperscript{2}
\\[1mm]
\textsuperscript{1}National Key Laboratory for Multimedia Information Processing, Peking University \\
\textsuperscript{2}Huawei Technologies Co., Ltd. \\
\small{
\texttt{\{lisida,liuyuanxin,linliyao\}@stu.pku.edu.cn} \quad
\texttt{\{lisc99,xusun\}@pku.edu.cn}
}\\
\small{
\texttt{\{baihaoli,lixiaohui33\}@huawei.com}
}
}

\begin{document}
\maketitle

\begin{abstract}
While reinforcement learning with verifiable reward (RLVR) enhances reasoning in large language models, its efficacy in long-horizon multimodal tasks remains underexplored. Visual Sudoku puzzles pose a representative challenge, requiring multi-step deductive reasoning over both local constraints and global dependency chains. To this end, we introduce FG-Sudoku, a fine-grained benchmark categorized by grid size and required deductive steps. Evaluations reveal that frontier MLLMs fail on complex long-range dependencies (e.g., 9×9 puzzles). During post-training, this near-zero success rate causes severe reward sparsity, halting standard RLVR. We propose Adaptive Curriculum Group Relative Policy Optimization (AdaC-GRPO) to overcome this. Unlike rigid or prior-based schedules, AdaC-GRPO dynamically constructs training batches by tracking the model's "Zone of Proximal Development" via online rollout accuracy. Experiments demonstrate that AdaC-GRPO enables Qwen2.5-VL-7B to approach the performance of Gemini-2.5-Flash while bridging 62.2\% of the overall accuracy gap to Gemini-2.5-Pro, with its generalizability further validated across InternVL-3.5 and Qwen3-VL. Project page: \url{https://github.com/DuHeng0128/AdaC-GRPO}.
\end{abstract}

\section{Introduction}
\label{sec:intro}
\begin{figure*}[t]
\centering
\includegraphics[width=1.0\linewidth]{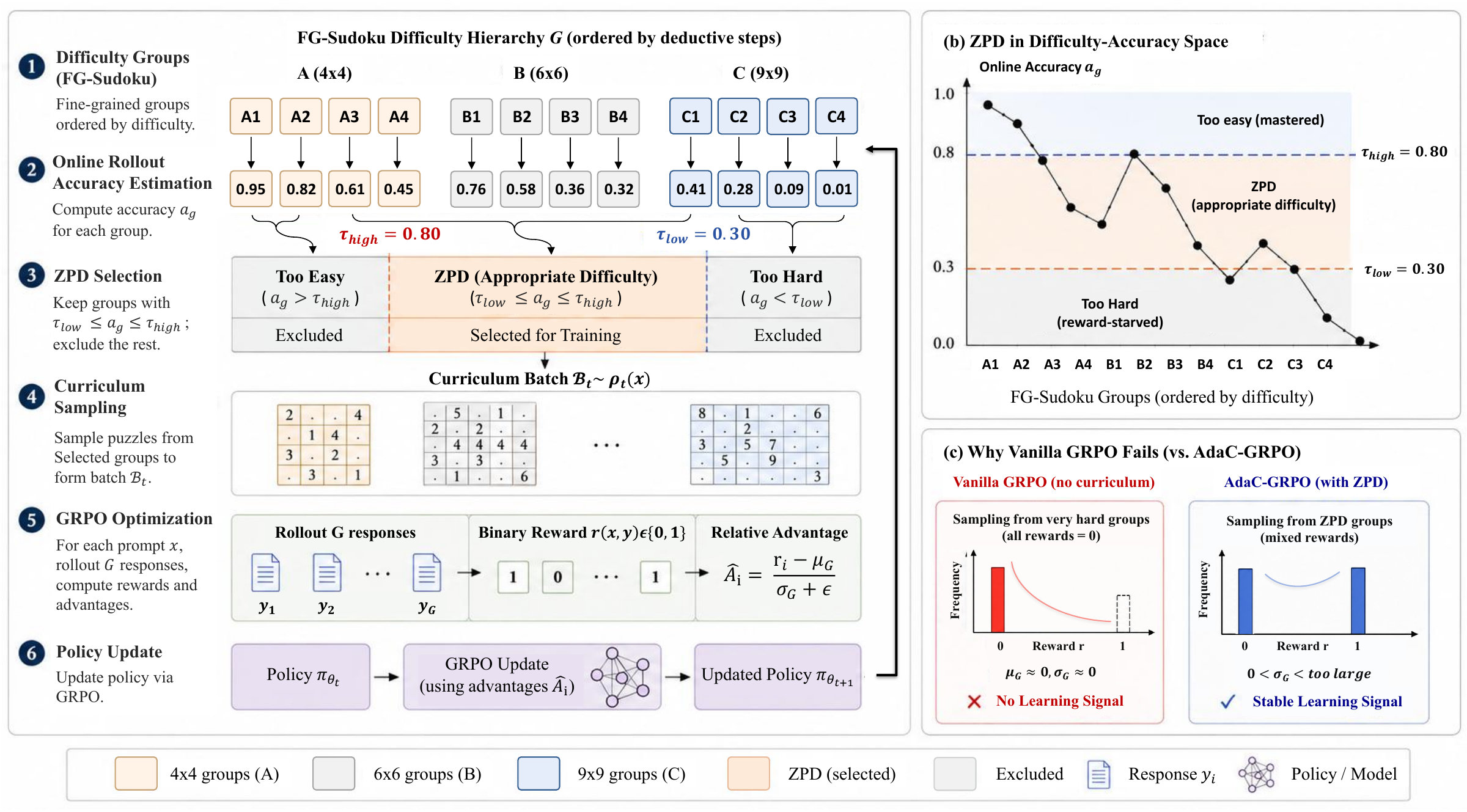}
\caption{Overview of AdaC-GRPO. The policy dynamically selects FG-Sudoku groups within the Zone of Proximal Development (ZPD) based on online rollout accuracy, excluding trivially solved and reward-starved tasks.}
\label{fig:algorithm}
\end{figure*}
Reinforcement learning with verifiable reward (RLVR) has achieved remarkable success in enabling large language models (LLMs) to tackle a wide range of reasoning tasks, including mathematical problem solving~\citep{Cobbe2021TrainingVT,Liu2024MathBenchET,Hendrycks2021MeasuringMP,Lu2023MathVistaEM}, general visual puzzles~\citep{chia2024puzzlevqa,Zhang2025PuzzleBenchAF}, and video reasoning~\citep{Hu2025VideoMMMUEK,Cheng2025VideoHolmesCM,Liu2025VideoReasonBenchCM,Qi2025VCRBenchAC}. However, while current paradigms excel at tasks requiring localized reasoning or pattern recognition, it remains unclear whether they can effectively handle \textbf{long-horizon multimodal reasoning} tasks, such as solving \textit{Visual Sudoku Puzzles}.

Sudoku is a logic-intensive puzzle that requires filling a partially completed $N\times N$ grid with numbers under strict constraints, involving multi-step deductive reasoning over both short-range local constraints and long-range global dependencies. The generalized Sudoku decision problem is NP-complete~\citep{yato2003complexity}, posing a formidable challenge for computational reasoning and serving as a clean abstraction of real-world constraint satisfaction problems.

In this work, we investigate the capability of multimodal large language models (MLLMs) to solve visual Sudoku puzzles (where the Sudoku grid is presented visually). Specifically, we seek to answer two key research questions: (1) To what extent can current MLLMs autonomously solve visual Sudoku puzzles without external symbolic solvers? (2) How can we improve their performance on such long-horizon reasoning tasks?

To address the first question, we introduce \textbf{FG-Sudoku}, a fine-grained diagnostic benchmark spanning multiple difficulty levels, chosen because Sudoku's difficulty is fully controllable by solver-derived deductive cost and its correctness is fully verifiable without any LLM-as-judge. The difficulty of each instance is quantified by the grid size and the number of deductive steps required by standard solving algorithms. Using this benchmark, we evaluate several frontier MLLMs and observe a clear limitation: while they perform competently on short-range constraints, their accuracy drops to near-zero on challenging 9$\times$9 instances due to failure in capturing long-range dependency chains. In RLVR post-training, this directly induces a \textbf{reward-variance collapse}: with a binary outcome reward and a near-zero pass rate, every sample in a GRPO group receives the same reward, the group-relative advantage degenerates to zero, and policy gradients vanish. This is exactly the regime where curriculum learning should help.

To answer the second question, we propose \textbf{Adaptive Curriculum Group Relative Policy Optimization (AdaC-GRPO)}. Existing curricula struggle with strict-constraint RL: static schedules (e.g., CurrReFT~\citep{deng2025curr}) advance regardless of the model's competence, while adaptive samplers (e.g., AdaRFT~\citep{Shi2025EfficientRF}) prioritize the highest-error instances. In long-horizon reasoning, however, the hardest instances yield a zero pass rate, so error-driven mining exacerbates reward starvation and halts optimization. AdaC-GRPO instead shifts from error-driven mining to \emph{competence-driven progression}: it uses online rollout accuracy not to find the hardest tasks but to pinpoint the model's "Zone of Proximal Development" (ZPD), bounding the active sampling distribution from both below and above, ensuring that each batch lies in the sweet spot where the model is sufficiently challenged yet still receives dense positive reinforcement.

Experimental results show that while standard supervised fine-tuning (SFT) followed by GRPO improves Qwen2.5-VL-7B on easy and medium Sudoku cases, the model continues to struggle on hard puzzles. In contrast, AdaC-GRPO lifts Qwen2.5-VL-7B's C-group accuracy from a near-zero baseline to 0.219, approaching Gemini-2.5-Flash on the hardest 9$\times$9 tier and closing 62.2\% of the overall gap to Gemini-2.5-Pro, while outperforming curriculum baselines such as AdaRFT and CurrReFT by over $5\times$ on the hardest tier. The framework also generalizes to InternVL-3.5 and Qwen3-VL.

Our main contributions are as follows:
\begin{itemize}
    \item We introduce \textbf{FG-Sudoku}, a diagnostic benchmark for evaluating MLLMs on visual Sudoku reasoning.
    \item We propose \textbf{AdaC-GRPO}, which addresses reward-variance collapse in binary-reward RLVR by constraining sampling to a Zone of Proximal Development. 
    \item We empirically show that AdaC-GRPO substantially improves Qwen2.5-VL-7B on visual Sudoku, consistently outperforming SFT, GRPO, CurrReFT, and AdaRFT on the same backbone. Our model approaches Gemini-2.5-Flash on the hardest 9$\times$9 tier, and closes 62.2\% of the overall gap to Gemini-2.5-Pro.
\end{itemize}


\section{Related Work}
\subsection{Symbolic and constraint reasoning}

Classical constraint satisfaction and symbolic reasoning frameworks form the foundation for combinatorial puzzles such as Sudoku~\citep{yato2003complexity}. Neuro-symbolic methods attempt to integrate differentiable learning with logic constraints~\citep{Bhuyan2024NeurosymbolicAI}, while recent hybrid systems enable models to propagate symbolic consistency through latent representations. For Sudoku, \citep{hathidara2023neuro} use a Neural Logic Machine with predefined predicates. However, these systems typically rely on explicit rule supervision or modular solvers, limiting scalability and flexibility. Our work takes a complementary approach: rather than embedding symbolic solvers into networks, we treat Sudoku as an end-to-end environment where the model must infer constraint structure implicitly through reward feedback.

\subsection{Multimodal reasoning and benchmarks}
Recent multimodal foundation models—including Gemini 2.5 \citep{comanici2025gemini}, Qwen2.5-VL \citep{Bai2025Qwen25VLTR}, GPT-4o \citep{hurst2024gpt}, DeepSeek-V3 \citep{liu2024deepseek}, and InternVL-3 \citep{zhu2025internvl3}—exhibit strong perception-language alignment yet continue to struggle with multi-step relational reasoning and constraint satisfaction. A growing body of benchmarks has sought to expose these weaknesses~\citep{luo2026mmmg}. ReasonBench \citep{Potamitis2025ReasonBENCHBT} and MATH-Bench \citep{Liu2024MathBenchET} emphasize symbolic and numerical reasoning; CLEVR \citep{Johnson2016CLEVRAD}, and GQA \citep{Hudson2019GQAAN} target visual compositionality and logical consistency; while ScienceQA \citep{Saikh2022ScienceQAAN} and VCR \citep{Zellers2018FromRT} explore multimodal question answering requiring factual grounding and inference. More recently, datasets such as ReasonVQA \citep{Tran2025ReasonVQAAM} extend this paradigm to visual abstraction and analogical reasoning, highlighting persistent failures of current models to maintain symbolic consistency across modalities. Meanwhile, reasoning-intensive puzzle benchmarks—such as SudokuBench \citep{Seely2025SudokuBenchEC}, Zebra-CoT \citep{Li2025ZebraCoTAD}, and PuzzleBench \citep{Zhang2025PuzzleBenchAF} further highlight the brittleness of current models under explicit symbolic constraints. Yet these datasets focus primarily on final-answer accuracy, leaving open the challenge of long-range constraint propagation.

Within this context, FG-Sudoku introduces a complementary perspective: a formally defined constraint-satisfaction reasoning task that preserves combinatorial complexity. It offers a continuous difficulty spectrum derived from solver statistics and enables quantitative analysis of end-to-end reasoning improvement. By focusing on a minimal but expressive domain, FG-Sudoku serves as a diagnostic counterpart to large-scale open-ended reasoning suites, providing a controlled environment for probing how multimodal models propagate local constraints into globally consistent solutions.

\subsection{Reinforcement fine-tuning and curriculum learning}

Reinforcement fine-tuning (RFT) has emerged as a central tool for improving reasoning behavior beyond supervised imitation~\citep{ouyang2022training,rafailov2023direct,christiano2017deep}. This paradigm has been prominently validated by recent frontier models achieving state-of-the-art reasoning capabilities, such as OpenAI o3 and the DeepSeek series (e.g., DeepSeek-V3 and R1)~\citep{liu2024deepseek, guo2025deepseek}. Recent extensions, including GRPO~\citep{Shao2024DeepSeekMathPT}, RRHF~\citep{yuan2023rrhf}, and DPO-based methods~\citep{rafailov2023direct}, reformulate the reward-weighted objective to improve stability. Multimodal variants such as RL-VLM-F~\citep{Wang2024RLVLMF} and Video-R1~\citep{Feng2025VideoR1RV} demonstrate that policy-gradient training can enhance factuality and consistency, but they remain highly vulnerable to sparse or discontinuous reward signals.

Classical work~\citep{Bengio2009CurriculumL} and recent adaptive curricula~\citep{Matiisen2017TeacherStudentCL,graves2017automated} dynamically adjust task difficulty to match a model’s evolving competence. In the reasoning domain, approaches such as AdaRFT~\citep{Shi2025EfficientRF}, CurrReFT~\citep{deng2025curr}, and GRPO~\citep{Shao2024DeepSeekMathPT} have improved optimization efficiency on sparse-reward tasks. Unlike static schedules or prior-based scoring that collapse on hard constraints, our proposed \textbf{AdaC-GRPO} dynamically tracks the model's \textit{Zone of Proximal Development (ZPD)} via online rollout accuracy. This structure-aware window ensures the dense positive feedback necessary for end-to-end optimization.

DAPO~\citep{yu2025dapo} also addresses zero-variance rollout groups, but it does so after sampling: prompts whose rewards are all zero or all one are discarded and replaced. AdaC-GRPO makes its routing decision before the training rollouts. It estimates competence at the difficulty-group level and assigns the next batch to a group inside the ZPD. The distinction matters in hard regions, where post-hoc filtering can spend most of the rollout budget before finding that a prompt provides no reward contrast for estimating group-relative advantages.


\begin{figure*}[t]
\centering
\includegraphics[width=\linewidth]{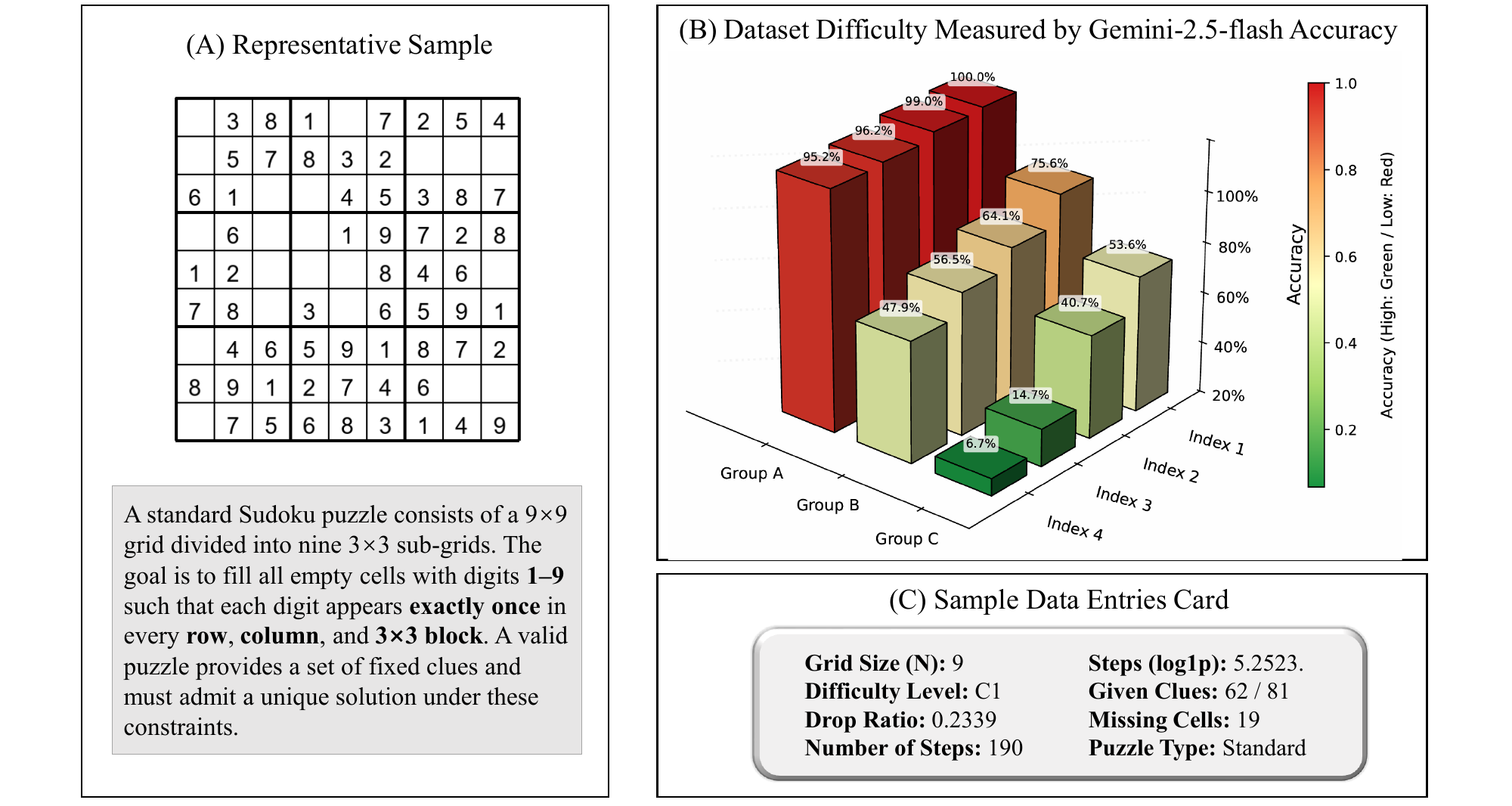}
\vspace{-1em}
\caption{Overview of the dataset, showing a representative Sudoku sample, model-measured difficulty steps across subsets, and the metadata structure of individual entries. Panel (B) reports Gemini-2.5-Flash's pass@1 accuracy on each subgroup as a post-hoc empirical validation of our solver-derived difficulty hierarchy; correctness is determined by exact-grid matching. Higher bars generally correspond to easier subgroups.}
\label{fig:dataset}
\vspace{-1em}
\end{figure*}

\section{FG-Sudoku: Dataset Construction}
\label{sec:dataset}
To systematically study long-horizon reasoning, we introduce FG-Sudoku, which contains puzzles at three canonical grid sizes: $4\times4$, $6\times6$, and $9\times9$. The complete dataset contains 3{,}072 training instances and 768 held-out test instances, uniformly distributed across grid sizes and carefully stratified into four difficulty tiers. We further use Gemini 2.5 Pro~\citep{comanici2025gemini} to synthesize high-quality \textbf{chain-of-thought (CoT) traces}. These traces can be used to bootstrap the model during the initial supervised fine-tuning (SFT) phase.

\subsection{Generation and Validation Pipeline.} 
For each size, we generate valid Sudoku boards and dynamically sample a drop ratio (fraction of given cells removed) to ensure layout diversity. Furthermore, we apply a deterministic constraint solver alongside a backtracking search (with an MRV heuristic) to eliminate instances with multiple solutions, guaranteeing strict unique solvability. The solver also records a structural trace of operations. By weighting and aggregating these operations, we derive a solver-based estimate of the deductive cost, $S$, for each instance. This highly optimized pipeline synthesizes 3K valid instances in roughly 60 seconds on a standard CPU, enabling flexible on-the-fly curriculum generation. See Appendix~\ref{sec:appendix_a} for detailed drop-ratio distributions and solver heuristics.

\subsection{Difficulty Stratification.} 
Driven by the $O(n^{n^2})$ search space, $S$ follows a heavily right-tailed distribution. Naive equal-width binning triggers \textit{curriculum starvation}: most instances concentrate in the easy bins, leaving the hard tail under-sampled and unable to produce a reliable training signal. We instead use \textbf{quantile-based partitioning} ($K{=}4$, balanced by count), which guarantees uniform sample mass across difficulty groups regardless of the distribution's skew. Formally, the $k$-th tier for grid size $n$ is defined by the empirical quantiles $q_k^{(n)}$ of $S$:
$$\mathcal{G}_{n, k} = \left\{ i \mid q_{k-1}^{(n)} \le s_i < q_k^{(n)} \right\}$$
This provides balanced data buffers across the continuous spectrum of deductive complexity, neutralizing the underlying heavy-tailed distribution.

\subsection{Advantages and Solver Traces.}
\begin{figure}[h]
\includegraphics[width=\linewidth]{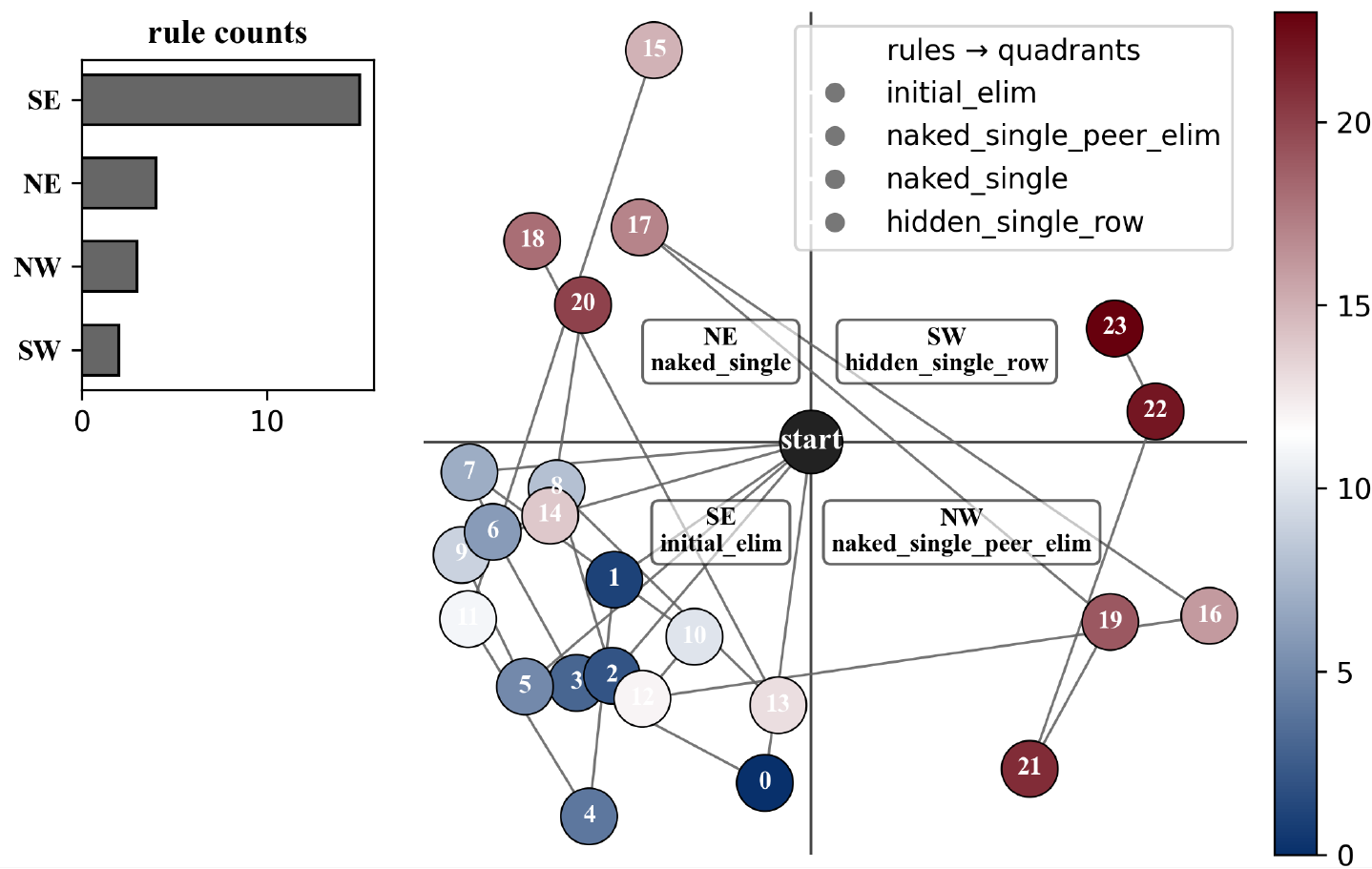}
\caption{Visualization of the solution trace for a 4×4 Sudoku puzzle. Each quadrant represents a distinct class of operations. Operations applied to the same cell are connected by straight-line links, and the colors change continuously from blue to red to indicate the temporal sequence of the reasoning steps.}
\label{fig:trace}
\vspace{-0.5em}
\end{figure}

Existing benchmarks such as SudokuBench and Zebra-CoT mainly provide curated puzzles for static evaluation, sometimes accompanied by natural-language reasoning. In FG-Sudoku, the recorded traces allow us to select instances by reasoning-path properties and compare model behavior under controlled settings. These properties include the structure of dependency chains and the number of steps over which global exclusivity constraints propagate.  Figure~\ref{fig:trace} shows a solution trace visualization of a $4\times4$ Sudoku puzzle, demonstrating that even seemingly simple instances require long, tightly coupled sequences of reasoning. 

\section{AdaC-GRPO}
\label{sec:AdaC-GRPO}
\subsection{Competence-Driven Curriculum via ZPD Scheduling}
Let $x \in \mathcal{X}$ be a Sudoku instance drawn from a time-varying task distribution $\rho_t(x)$. Due to the combinatorial state space $\mathcal{O}(n^{n^2})$, uniform sampling across the full difficulty hierarchy $\mathcal{G}$ produces highly sparse reward signals and uninformative policy gradients on complex instances.
 
AdaC-GRPO reframes the curriculum as a dynamic distribution-matching problem bounded by the model's \textit{Zone of Proximal Development} (ZPD). At curriculum epoch $t$ with current policy $\pi_{\theta_t}$, we define the empirical rollout accuracy for a difficulty group $g \in \mathcal{G}$ as
\begin{equation}
\small
    a_g^{(t)} = \hat{\mathbb{E}}_{\substack{x \sim \mathcal{D}_g \\ \mathbf{y} \sim \pi_{\theta_t}(\cdot \mid x)}} \! \Big[ \mathbf{1}\!\big( \text{Grid}(\mathbf{y}) = \mathbf{y}^\star_{\text{grid}}(x) \big) \Big],
\end{equation}
where $\mathcal{D}_g$ is the uniform distribution over instances in group $g$, $\mathbf{y}^\star_{\text{grid}}(x)$ is the ground-truth solution grid, and $\text{Grid}(\mathbf{y})$ extracts the answer matrix from the model output. The indicator checks \textbf{exact cell-by-cell match}. The ZPD is the feasibility region $[\tau_{\text{low}}, \tau_{\text{high}}]$ that filters out both trivially mastered and overwhelmingly complex tasks. We schedule the training target $g^*_t$ by
\begin{equation}
    g^*_t = \arg\max_{g \in \mathcal{G}} \Big( a_g^{(t)} \cdot \mathbb{I}_{\text{ZPD}}(g, t) \Big),
\end{equation}
where $\mathbb{I}_{\text{ZPD}}(g, t) = \mathbf{1}(\tau_{\text{low}} \le a_g^{(t)} \le \tau_{\text{high}})$. The active sampling distribution $\rho_t(x)$ is then deterministically routed to $\mathcal{D}_{g^*_t}$, keeping training within a dense reward landscape. The selection rule additionally handles cases where no candidate group lies within the ZPD. If all candidate groups satisfy $a_g^{(t)} > \tau_{\text{high}}$, training terminates since the curriculum has been fully mastered. Otherwise, the policy rolls back to the previously trained group $g^*_{t-1}$ for additional consolidation, allowing the model to re-enter the ZPD before further progression.

The Sudoku solver is used only to construct the static groups $\mathcal{D}_g$ and their reference solutions. AdaC-GRPO does not query the solver during training. The scheduler uses the policy's online rollout accuracy $a_g^{(t)}$, so other tasks can define groups from structural metadata, human labels, or empirical pass rates.
 
\subsection{Group Relative Policy Optimization for Strict Constraints}
Constraint satisfaction is brittle: a single invalid assignment corrupts the global solution. We therefore use an outcome-based binary reward complemented by an auxiliary format term:
\begin{equation}
\small
    r(x, \mathbf{y}) = \underbrace{\mathbf{1}\!\left[\text{Grid}(\mathbf{y}) {=} \mathbf{y}^\star_{\text{grid}}(x)\right]}_{r_{\text{acc}}(x,\mathbf{y})} \, + \, \lambda \cdot r_{\text{format}}(\mathbf{y}),
\end{equation}
where $r_{\text{format}}(\mathbf{y}) = |\{c \in \mathbf{y} : c \in \mathcal{P}\}| / |\mathbf{y}|$. $\mathcal{P}$ denotes the set of tokens conforming to the predefined parsable output format. The combination $r(x, \mathbf{y})$ drives the GRPO advantage in Eq.~(4). However, the format term $r_{\text{format}}$ is not used for ZPD selection, as we are more concerned with changes in problem-solving ability. We also explored lenient, dense intermediate rewards (e.g., partial assignment accuracy), but found they emit false-positive advantage signals that reinforce trajectories satisfying local but violating global constraints.
 
We optimize the policy with the standard GRPO framework~\citep{Shao2024DeepSeekMathPT}. For each prompt $x \sim \rho_t(x)$, we sample a group of $G$ outputs $\{\mathbf{y}_1, \dots, \mathbf{y}_G\} \sim \pi_{\theta_{\text{old}}}(\cdot \mid x)$ and compute the group-relative advantage
\begin{equation}
    \hat{A}_i = \frac{r(x, \mathbf{y}_i) - \mu_G}{\sigma_G + \epsilon},
\end{equation}
where $\mu_G$ and $\sigma_G$ are the mean and standard deviation of the group rewards.
 
\textbf{Why impose a lower bound $\tau_{\text{low}}$?}
Error-driven curricula such as AdaRFT~\citep{Shi2025EfficientRF} treat higher error as greater learning potential and route training toward the hardest instances. In binary-reward RLVR, sufficiently hard prompts often yield $r_{\text{acc}}=0$ for all $G$ rollouts, giving $\mu_G=\sigma_G=0$ and $\hat{A}_i\equiv0$. Such groups provide no task-success learning signal. DAPO~\citep{yu2025dapo} handles this by oversampling and retaining only groups containing both successful and failed rollouts. Its post-hoc filter ensures nonzero variance for retained groups, although discarded groups still consume rollouts. For a prompt with independent rollout success probability $p$, the probability of obtaining an informative group is
\begin{equation}
q(p)=1-(1-p)^G-p^G.
\end{equation}
The expected number of raw rollouts required to obtain one retained group is therefore
\begin{equation}
\mathbb{E}[N_{\mathrm{roll}}]
=\frac{G}{q(p)}
\approx\frac{1}{p},
\qquad Gp\ll1.
\end{equation}
This cost grows rapidly as $p$ approaches zero. AdaC-GRPO uses online competence estimates to restrict $\rho_t(x)$ before generation, postponing groups whose estimated rollout accuracy falls below $\tau_{\text{low}}$ until their accuracy rises into the ZPD as the policy improves, thereby keeping the success probability of sampled tasks away from zero. If exactly one of the $G$ rollouts succeeds, that trajectory is the sole source of positive credit, and its normalized advantage scales as $\mathcal{O}(\sqrt{G})$ ($\sqrt{G-1}$ under the population-standard-deviation convention). Since final-answer rewards do not assess intermediate reasoning, the update may reinforce a success that the model cannot reliably reproduce. Restricting training to the ZPD reduces dependence on such rare events.

\section{Experiments}
\label{sec:experiments}

\begin{table*}[!h]
\centering
\subfloat[Detailed per-category results.]{
    \label{tab:detailed}
    \resizebox{\textwidth}{!}{
        \begin{tabular}{lcccc|cccc|cccc}
        \toprule
        Model & A1 & A2 & A3 & A4 & B1 & B2 & B3 & B4 & C1 & C2 & C3 & C4\\
        \midrule
        \addlinespace[4pt]
        \rowcolor{groupA}\multicolumn{13}{l}{\textbf{Block A: Proprietary Models}}\\
        GPT-4o~\citep{hurst2024gpt} & 0.729 & 0.598 & 0.377 & 0.571 & 0.171 & 0.094 & 0.048 & 0.042 & 0.000 & 0.000 & 0.000 & 0.000 \\
        \rowcolor{groupA!60} o3~\citep{openai_o3_2025} & \uline{1.000} & 0.979 & 0.962 & \uline{1.000} & \uline{0.939} & \uline{0.891} & 0.855 & \uline{0.938} & 0.203 & 0.136 & 0.059 & 0.000 \\
        Claude-Sonnet-4~\citep{claudesonnet2025} & 0.965 & 0.907 & 0.830 & 0.619 & 0.829 & 0.781 & 0.565 & 0.708 & \uline{0.754} & 0.407 & 0.206 & 0.017 \\
        \rowcolor{groupA!60} DeepSeek-V3~\citep{liu2024deepseek} & 0.565 & 0.464 & 0.547 & 0.429 & 0.268 & 0.281 & 0.177 & 0.188 & 0.072 & 0.017 & 0.000 & 0.000 \\
        Gemini-2.5-Flash~\citep{comanici2025gemini} & \uline{1.000} & 0.990 & 0.962 & 0.952 & 0.756 & 0.641 & 0.565 & 0.479 & 0.536 & 0.407 & 0.147 & 0.067 \\
        \rowcolor{groupA!60} Gemini-2.5-Pro~\citep{comanici2025gemini} & 0.988 & \uline{1.000} & \uline{0.981} & \uline{1.000} & \uline{0.939} & 0.859 & \uline{0.935} & 0.917 & 0.681 & \uline{0.814} & \uline{0.735} & \uline{0.383} \\
        \midrule
        \addlinespace[4pt]
        \rowcolor{groupB}\multicolumn{13}{l}{\textbf{Block B: Open-weight Models}}\\
        LLaVA-NeXT-34B~\citep{liu2024llavanext} & 0.024 & 0.000 & 0.000 & 0.000 & 0.000 & 0.000 & 0.000 & 0.000 & 0.000 & 0.000 & 0.000 & 0.000 \\
        \rowcolor{groupB!60}Video-R1~\citep{Feng2025VideoR1RV} & 0.282 & 0.093 & 0.113 & 0.000 & 0.000 & 0.000 & 0.000 & 0.000 & 0.000 & 0.000 & 0.000 & 0.000 \\
        InternVL3-8B~\citep{zhu2025internvl3} & 0.008 & 0.006 & 0.002 & 0.005 & 0.000 & 0.000 & 0.000 & 0.000 & 0.000 & 0.000 & 0.000 & 0.000 \\
        \rowcolor{groupB!60}InternVL3.5-8B~\citep{wang2025internvl3} & 0.765 & 0.588 & 0.453 & 0.333 & 0.220 & 0.062 & 0.048 & 0.021 & 0.029 & 0.000 & 0.000 & 0.000 \\
        Qwen2.5-VL-7B~\citep{Bai2025Qwen25VLTR} & 0.624 & 0.361 & 0.264 & 0.143 & 0.061 & 0.016 & 0.000 & 0.000 & 0.000 & 0.000 & 0.000 & 0.000 \\
        \rowcolor{groupB!60} Qwen3-VL-8B~\citep{bai2025qwen3} & 0.906 & 0.784 & 0.642 & 0.476 & 0.634 & 0.406 & 0.177 & 0.104 & 0.058 & 0.051 & 0.000 & 0.000 \\
        \midrule
        \addlinespace[4pt]
        \rowcolor{groupC}\multicolumn{13}{l}{\textbf{Block C: Training Paradigms}}\\
        Qwen2.5-VL-7B + SFT & 0.624 & 0.495 & 0.340 & 0.333 & 0.159 & 0.031 & 0.032 & 0.021 & 0.058 & 0.034 & 0.000 & 0.000 \\
        \rowcolor{groupC!60} Qwen2.5-VL-7B + GRPO~\citep{Shao2024DeepSeekMathPT} & 0.906 & 0.835 & 0.509 & 0.476 & 0.732 & 0.422 & 0.355 & 0.292 & 0.087 & 0.068 & 0.000 & 0.000 \\
        Qwen2.5-VL-7B + CurrReFT~\citep{deng2025curr} & 0.588 & 0.722 & 0.623 & 0.619 & 0.793 & 0.516 & 0.468 & 0.312 & 0.043 & 0.000 & 0.000 & 0.000 \\
        \rowcolor{groupC!60} Qwen2.5-VL-7B + AdaRFT~\citep{Shi2025EfficientRF} & 0.788 & 0.784 & 0.698 & 0.571 & 0.646 & 0.531 & 0.500 & 0.333 & 0.116 & 0.051 & 0.000 & 0.000 \\
        Qwen2.5-VL-7B + DAPO~\citep{yu2025dapo} & \textbf{0.976} & \textbf{0.948} & \textbf{0.792} & 0.619 & 0.866 & 0.609 & 0.516 & \textbf{0.438} & 0.232 & 0.102 & 0.000 & 0.000 \\
        \rowcolor{groupC!60} Qwen2.5-VL-7B + AdaC-GRPO \textbf{(Ours)} & \textbf{0.976} & 0.928 & 0.774 & \textbf{0.667} & \textbf{0.878} & \textbf{0.641} & \textbf{0.548} & 0.396 & \textbf{0.478} & \textbf{0.288} & \textbf{0.074} & \textbf{0.017} \\
        \midrule
        \addlinespace[4pt]
        \rowcolor{groupD}\multicolumn{13}{l}{\textbf{Block D: Architecture Generalization}}\\
        InternVL3.5-8B + AdaC-GRPO \textbf{(Ours)} & 0.976 & 0.897 & 0.792 & 0.714 & 0.854 & 0.609 & 0.581 & 0.417 & 0.420 & 0.322 & 0.059 & 0.033 \\
        \rowcolor{groupD!60} Qwen3-VL-8B + AdaC-GRPO \textbf{(Ours)} & 0.965 & 0.938 & 0.830 & 0.667 & 0.902 & 0.656 & 0.548 & 0.354 & 0.507 & 0.339 & 0.103 & 0.067 \\
        \bottomrule
        \end{tabular}
    }
}
\par
\noindent
\subfloat[Aggregated group-level results.]{
    \label{tab:aggregate_1}
    \resizebox{0.48\textwidth}{!}{
        \begin{tabular}[t]{lccc|c}
        \toprule
        Model & A & B & C & Total \\
        \midrule
        \rowcolor{groupA}\multicolumn{5}{l}{\textbf{Block A: Proprietary Models}}\\
        GPT-4o & 0.5935 & 0.0978 & 0.0000 & 0.2304 \\
        \rowcolor{groupA!60}o3 & 0.9842  & 0.9065  & 0.1017  & 0.6641  \\
        Claude-Sonnet-4 & 0.8867  & 0.7304 & 0.3557 & 0.6576 \\
        \rowcolor{groupA!60}DeepSeek-V3 & 0.5118  & 0.2342  & 0.0233  & 0.2565 \\
        Gemini-2.5-Flash & 0.9844  & 0.6291  & 0.2930  & 0.6355  \\
        \rowcolor{groupA!60}Gemini-2.5-Pro & \uline{0.9921}  & \uline{0.9139}  & \uline{0.6562}  & \uline{0.8540} \\
        \midrule
        \addlinespace[4pt]
        \rowcolor{groupC}\multicolumn{5}{l}{\textbf{Block C: Training Paradigms}}\\
        Qwen2.5-VL-7B + SFT & 0.4925  & 0.0704  & 0.0235  & 0.1954  \\
        \rowcolor{groupC!60}Qwen2.5-VL-7B + GRPO & 0.7616 & 0.4807 & 0.0391 & 0.4271 \\
        Qwen2.5-VL-7B + CurrReFT & 0.6486  & 0.5549  & 0.0116  & 0.4050  \\
        \rowcolor{groupC!60}Qwen2.5-VL-7B + AdaRFT & 0.7501  & 0.5232  & 0.0430  & 0.4388 \\
        Qwen2.5-VL-7B + DAPO & \textbf{0.8984} & 0.6367 & 0.0859 & 0.5404 \\
        \rowcolor{groupC!60}
        Qwen2.5-VL-7B + AdaC-GRPO & 0.8906 & \textbf{0.6485} & \textbf{0.2189} & \textbf{0.5860} \\
        \bottomrule
        \end{tabular}
    }
}%
\hfill
\subfloat[Aggregated group-level results.]{
    \label{tab:aggregate_2}
    \resizebox{0.48\textwidth}{!}{
    \begin{tabular}[t]{lccc|c}
    \toprule
    Model & A & B & C & Total \\
    \midrule
    \addlinespace[4pt]
    \rowcolor{groupB}\multicolumn{5}{l}{\textbf{Block B: Open-weight Models}}\\
    LLaVA-NeXT-34B & 0.0080 & 0.0000 & 0.0000 & 0.0027\\
    \rowcolor{groupB!60}Video-R1 & 0.1523 & 0.0000 & 0.0000 & 0.0508  \\
    InternVL3-8B & 0.0117  & 0.0000  & 0.0000  & 0.0039  \\
    \rowcolor{groupB!60}InternVL3.5-8B & 0.5979 & 0.1015 & 0.0078  & 0.2357  \\
    Qwen2.5-VL-7B & 0.4104  & 0.0235  & 0.0000  & 0.1446  \\
    \rowcolor{groupB!60}Qwen3-VL-8B & 0.7698  & 0.3669 & 0.0274 & 0.3881  \\
    \midrule
    \addlinespace[4pt]
    \rowcolor{groupD}\multicolumn{5}{l}{\textbf{Block D: Architecture Generalization}}\\
    InternVL3.5-8B + AdaC-GRPO & 0.8865 & 0.6447 & 0.2108 & 0.5807 \\
    \rowcolor{groupD!60}Qwen3-VL-8B + AdaC-GRPO & \textbf{0.9024} & \textbf{0.6520} & \textbf{0.2578} & \textbf{0.6041} \\
    \bottomrule
    \end{tabular}
    }
}
\caption{Benchmarking leading proprietary and open-weight multimodal models (Pass@1). Panel (a) reports per-tier results, while panels (b) and (c) report group-level averages. A, B, and C denote the 4$\times$4, 6$\times$6, and 9$\times$9 subsets, respectively; 1--4 denote difficulty tiers. Block C compares training methods under the same Qwen2.5-VL-7B backbone, SFT warm start, reward, and rollout budget. \textbf{Bold} denotes the best result within Block C, while \underline{underlined} values indicate the best result across all models.}
\vspace{-1.0em}
\end{table*}

We structure our evaluation around four core research questions: (\textbf{RQ1}) Can FG-Sudoku accurately distinguish the reasoning capabilities of multimodal models across varying grid sizes and structural difficulties? (\textbf{RQ2}) Does AdaC-GRPO improve sample efficiency and final performance over vanilla GRPO? (\textbf{RQ3}) Is fine-grained difficulty stratification strictly necessary? (\textbf{RQ4}) Can trace alignment provide actionable diagnostics?

\subsection{Experimental Setup}
\label{subsec:setup}

\paragraph{Baselines and Metrics.} 
We evaluate a diverse set of frontier multimodal large language models (MLLMs), including proprietary systems (e.g., Gemini 2.5 Pro, GPT-4o) and open-weight alternatives (e.g., Qwen2.5-VL, InternVL3.5). The prompts used in the evaluation are listed in Appendix~\ref{test_prompt}. We report \textit{Pass@1} as the primary metric.

\paragraph{AdaC-GRPO Implementation Details.} 
We implement AdaC-GRPO on top of the open-source training codebase released by Video-R1~\citep{Feng2025VideoR1RV}, which provides multimodal GRPO infrastructure (the Video-R1 fine-tuned \emph{model checkpoint} is reported separately as a baseline in Table~\ref{tab:detailed}). Prior to RL optimization, we cold-start the model's step-by-step reasoning format and rule adherence by supervised fine-tuning (SFT) on 300 Gemini-2.5-Pro-annotated traces. During AdaC-GRPO training, we sample $G=8$ candidate completions per prompt, utilizing the combined binary accuracy and formatting reward defined in Section~\ref{sec:AdaC-GRPO}. Our curriculum schedule dynamically routes the active sampling distribution strictly to difficulty groups whose validation accuracy falls within the empirical ZPD bounds $[\tau_{\text{low}}, \tau_{\text{high}}] = [0.3, 0.8]$. We additionally provide threshold sensitivity and selection-rule ablations in Appendix~\ref{appendix:details}, to qualitatively determine the upper and lower bounds and the argmax selection mechanism; full hyperparameter configurations are also detailed there.

\subsection{Main Results (RQ1 \& RQ2)}
\label{subsec:main_results}

The benchmark results (Table~\ref{tab:detailed}) reveal a clear long-horizon reasoning bottleneck shared across the frontier. Addressing \textbf{RQ1}, while most frontier models perform competently on local 4$\times$4 grids (Group A), even strongly capable models collapse to near-zero on the combinatorial 9$\times$9 Group C. This sharp degradation as deductive cost grows isolates multi-step constraint propagation, rather than perception, as the limiting factor.

Addressing \textbf{RQ2}, AdaC-GRPO raises overall Pass@1 from 0.427 with GRPO to 0.586. DAPO provides a stronger baseline than vanilla GRPO: its dynamic sampling reaches 0.540 overall, but only 0.086 on Group C, compared with 0.219 for AdaC-GRPO. A rollout-efficiency analysis further shows that, upon entering Group C, 99.7\% of DAPO rollout groups have zero accuracy-reward variance, compared with 54.7\% for AdaC-GRPO. This shows that curriculum progression yields more informative rollouts when direct sampling rarely succeeds (Appendix~\ref{appendix:rollout_efficiency}). AdaC-GRPO also outperforms curriculum-based methods such as CurrReFT and AdaRFT at every difficulty level, with the gap widening substantially on harder instances. Appendix~\ref{appendix:seeds} reports variance across runs.

\paragraph{Disentangling the SFT warm-start.}
To isolate the contribution of the 300 Gemini-generated SFT traces from the RL gain, the \texttt{+SFT} row in Table~\ref{tab:detailed} and~\ref{tab:aggregate_1} serve as the warm-start-only baseline. SFT alone achieves $0.023$ on C-group --- essentially identical to the untuned Qwen2.5-VL-7B baseline ($0.000$) and far below AdaC-GRPO ($0.219$). The hard-instance gain therefore cannot be attributed to teacher-generated CoT data; it stems from the curriculum-bounded RL phase.

\paragraph{Closing the gap to proprietary models.}
On Block C of Table~\ref{tab:aggregate_1}, AdaC-GRPO closes 62.2\% of the overall accuracy gap between the untuned Qwen2.5-VL-7B baseline and Gemini-2.5-Pro (and 33.4\% on C-group). On the hardest 9$\times$9 tier specifically, AdaC-GRPO (0.219) approaches Gemini-2.5-Flash (0.293). This indicates that competence-driven ZPD scheduling can lift a 7B open-weight model toward mid-tier proprietary performance on long-horizon constraint reasoning.

\subsection{Curriculum Trajectory}
\label{subsec:unit_testing}

\begin{figure}[h]
\centering
\includegraphics[width=\linewidth]{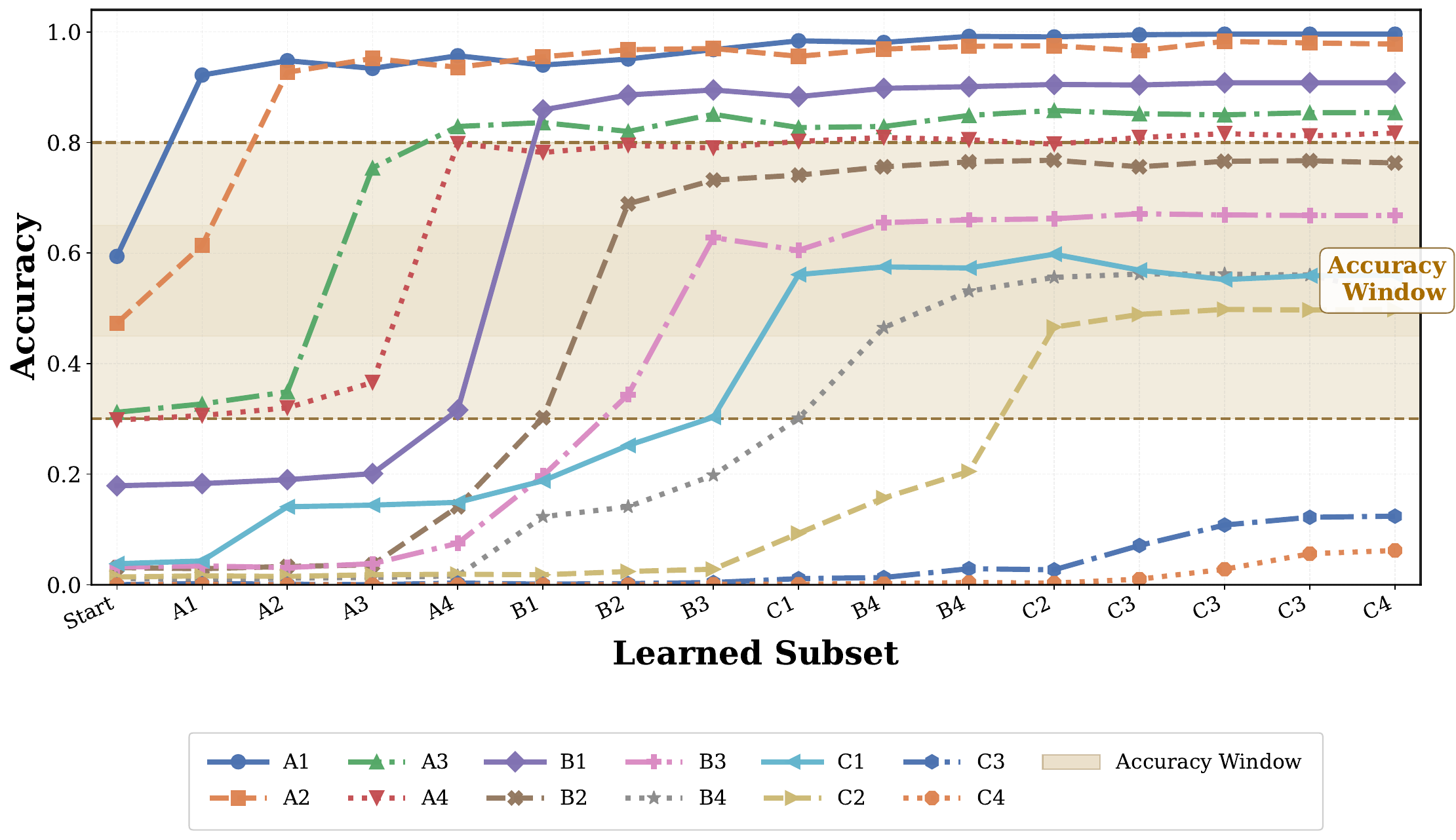}
\caption{Trajectory of policy competence during curriculum-guided unit testing. The horizontal axis represents sequentially activated training subsets, while the vertical axis shows the rollout accuracy ($f=1/4$). The orange shaded area denotes the globally defined Zone of Proximal Development (ZPD) bounds $[\tau_{\text{low}}, \tau_{\text{high}}]$.}
\label{fig:unit}
\vspace{-1em}
\end{figure}

To understand how AdaC-GRPO scales the difficulty hierarchy, we monitor the online rollout accuracy on held-out validation subsets across training epochs (Figure~\ref{fig:unit}). The trajectory shows a clear sequential dependency between subsets. By structurally grouping instances and dynamically activating subsets based on the $[\tau_{\text{low}}, \tau_{\text{high}}]$ threshold, AdaC-GRPO keeps the active policy within its ZPD (orange shaded region), supplying the dense gradient signals required to master increasingly complex global dependencies.

\subsection{Ablation and Analysis}
\label{sec:ablation}
\begin{table}[h]
\centering
\small
\textbf{(a) Impact of Curriculum Granularity} \\
\vspace{1mm}
\resizebox{0.49\textwidth}{!}{\begin{tabular}{lccc|c}
\toprule
Curriculum Granularity & A & B & C & Total\\
\midrule
Coarse-grained (size only) & 0.6484  & 0.5547  & 0.0117  & 0.4049 \\
\rowcolor{gray!20} Fine-grained (ours) & \textbf{0.8906} & \textbf{0.6485} & \textbf{0.2189} & \textbf{0.5860} \\
\bottomrule
\end{tabular}
}
\vspace{4mm}

\textbf{(b) Impact of Reward Formulation} \\
\vspace{1mm}
\resizebox{0.49\textwidth}{!}{\begin{tabular}{lccc|c}
\toprule
Reward Formulation & A & B & C & Total\\
\midrule
Lenient (partial) Reward & 0.5860 & 0.1250 & 0.0000 & 0.2370 \\
\rowcolor{gray!20} Binary Reward (ours) & \textbf{0.8906} & \textbf{0.6485} & \textbf{0.2189} & \textbf{0.5860} \\
\bottomrule
\end{tabular}
}
\vspace{4mm}

\textbf{(c) Train/Evaluation Modality} \\
\vspace{1mm}
\resizebox{0.49\textwidth}{!}{\begin{tabular}{lccc|c}
\toprule
Train $\rightarrow$ Eval & A & B & C & Total \\
\midrule
Untrained $\rightarrow$ Text & 0.535 & 0.047 & 0.004 & 0.195 \\
Untrained $\rightarrow$ Visual & 0.410 & 0.024 & 0.000 & 0.145 \\
Text $\rightarrow$ Text & \textbf{0.906} & 0.582 & 0.125 & 0.538 \\
\rowcolor{gray!20}Visual $\rightarrow$ Visual (ours) & 0.891 & \textbf{0.649} & \textbf{0.219} & \textbf{0.586} \\
Text $\rightarrow$ Visual & 0.707 & 0.496 & 0.082 & 0.428 \\
Visual $\rightarrow$ Text & 0.859 & 0.547 & 0.109 & 0.505 \\
\bottomrule
\end{tabular}
}
\caption{Ablation results. \textbf{(a)} Fine-grained tiers bridge the gap between grid sizes. \textbf{(b)} Partial reward fails on Group C. \textbf{(c)} Each trained condition uses one input form throughout training; cross-modal rows change only the evaluation input.}
\vspace{-1mm}
\label{tab:combined_ablations}
\end{table}

\paragraph{Necessity of Fine-Grained Stratification (RQ3).} 
To address \textbf{RQ3}, we ablate the fine-grained quantile stratification by partitioning instances solely by grid size (4$\times$4, 6$\times$6, 9$\times$9). As shown in Table~\ref{tab:combined_ablations}(a), this coarse-grained curriculum degrades performance, particularly on the C-group ($0.219 \to 0.012$). Without fine-grained stepping stones, the structural difficulty gap between 6$\times$6 and 9$\times$9 grids is too large to bridge in one step; the model suffers reward starvation, failing to maintain a viable ZPD and halting optimization prematurely.

\paragraph{Binary vs.\ lenient rewards.} 
In Section~\ref{sec:AdaC-GRPO}, we argued that lenient intermediate rewards, such as partial-grid accuracy, can assign positive learning signals to trajectories that are locally plausible but globally invalid. Table~\ref{tab:combined_ablations}(b) supports this hypothesis: an AdaC-GRPO variant trained with partial rewards performs well on the simple A-group instances but drops to $0.000$ on the C-group. In globally constrained tasks such as Sudoku, local correctness does not necessarily imply progress toward a globally feasible solution. Binary outcome rewards therefore better align the optimization signal with exact task success and avoid reinforcing partially correct but ultimately invalid trajectories, particularly when successful rollouts are rare on difficult instances.

\paragraph{Visual vs.\ textual input modality.}
Table~\ref{tab:combined_ablations}(c) disentangles training and evaluation modality. Without training, text input is consistently easier than the rendered board. Under matched train--evaluation modalities, visual training performs better overall and on Groups B and C, while text remains stronger on Group A. Cross-modal evaluation also shows stronger transfer from visual to text, suggesting that the two-dimensional layout provides additional benefit as constraint reasoning becomes more difficult.

\paragraph{Trace-Based Diagnostics (RQ4).}
\label{subsec:error_analysis}
Trace alignment shows that the baselines often lose track of the puzzle state and introduce duplicate digits within a row, column, or box. Relative to the SFT baseline on the same evaluated traces, AdaC-GRPO reduces the count of these solver-detected constraint conflicts by 67.4\% (Each duplicate pair counts as one conflict). Its remaining errors are more often associated with higher-order constraint patterns such as \textit{Hidden Pairs}. Appendix~\ref{appendix:Error} gives the violation taxonomy and examples.

\paragraph{Generalization across tasks.}
\begin{table}[t]
\centering
\resizebox{0.45\textwidth}{!}{\begin{tabular}{@{}lcccc@{}}
\toprule
Method & Easy & Medium & Hard & Overall \\
\midrule
\rowcolor{gray!20}\multicolumn{5}{@{}l}{\textit{ZebraLogicBench-ID}~\citep{lin2025zebralogic}} \\
Base & 0.453 & 0.047 & 0.014 & 0.165 \\
GRPO & \textbf{0.656} & 0.063 & 0.014 & 0.235 \\
AdaC-GRPO & 0.625 & \textbf{0.109} & \textbf{0.042} & \textbf{0.250} \\
\addlinespace[2pt]
\rowcolor{gray!20}\multicolumn{5}{@{}l}{\textit{CMM-Math}~\citep{cmmmath}} \\
Base & 0.526 & 0.480 & 0.410 & 0.469 \\
GRPO & 0.537 & 0.493 & 0.416 & 0.477 \\
AdaC-GRPO & \textbf{0.540} & \textbf{0.508} & \textbf{0.427} & \textbf{0.487} \\
\bottomrule
\end{tabular}}
\caption{Pass@1 after training GRPO or AdaC-GRPO from the same task-specific initialization. ZebraLogicBench uses structural-complexity tiers; CMM-Math uses its provided level labels.}
\label{tab:method_generalization}
\vspace{-0.5em}
\end{table}
We train AdaC-GRPO and GRPO separately on each benchmark from the same task-specific initialization. AdaC-GRPO improves performance on the medium and hard tiers of both tasks. ZebraLogicBench contains only 64--72 examples in each of these tiers, so the absolute differences should be interpreted cautiously. CMM-Math contains roughly 1,800--2,200 examples per tier and exhibits the same trend with substantially larger evaluation sets. These results evaluate the task-level generalization of the scheduling strategy beyond Sudoku. Additionally, Appendix~\ref{appendix:transfer} evaluates zero-shot checkpoint transfer from the Sudoku-trained model to out-of-domain tasks and shows that AdaC-GRPO reduces the average reasoning-token usage by roughly 50\%.

\section{Conclusion}
\label{sec:conclusion}
This work addresses the reward starvation challenge in applying RL to long-horizon visual reasoning. We introduce FG-Sudoku, a diagnostic benchmark isolating multi-step constraint bottlenecks. To overcome optimization collapse on complex tasks, we propose AdaC-GRPO, a competence-driven curriculum that confines active sampling to the model’s Zone of Proximal Development (ZPD). Experiments show AdaC-GRPO stabilizes gradient signals and lifts Qwen2.5-VL-7B’s C-group accuracy from a near-zero baseline to 0.219, substantially outperforming standard GRPO and existing curriculum-learning methods. It approaches Gemini-2.5-Flash on the hardest tier and closes 62.2\% of the overall gap to Gemini-2.5-Pro. Future work will explore broader task transfer and principled multi-task curricula for real-world long-horizon reasoning problems.

\section*{Limitations}
\noindent\textbf{Compute and statistical coverage.} We repeat GRPO and AdaC-GRPO across three training seeds, while the remaining training baselines are reported from single runs due to the high computational cost of multimodal RL. The hard tiers of ZebraLogicBench also contain relatively few test examples, leading to higher sampling uncertainty in the reported differences.

\noindent\textbf{Difficulty partitions.} AdaC-GRPO assumes access to a verifiable reward and a meaningful partition of the task space. Such partitions may be difficult to construct in open-ended domains where reliable metadata or stable empirical difficulty tiers are unavailable.

\noindent\textbf{Scope of hard-case and cross-task evaluation.} Performance remains low on the hardest C3 and C4 tiers. FG-Sudoku currently covers standard 4$\times$4, 6$\times$6, and 9$\times$9 puzzles, without evaluating irregular variants or larger grids. Beyond Sudoku, our experiments include one text-based CSP benchmark and one multimodal mathematics benchmark; broader validation on open-ended visual reasoning tasks remains for future work.

\section*{Acknowledgments}

We thank the anonymous reviewers for their constructive feedback. Xu Sun is the corresponding author of this paper.

\bibliography{custom}
\appendix
\section{Detailed Dataset Construction}
\label{sec:appendix_a}

\subsection{Drop-Ratio Sampling Distributions}
To control the initial hardness of the generated puzzles while avoiding trivially unsatisfiable or entirely degenerate layouts, we dynamically sample the \textit{drop ratio} (the fraction of given cells removed from a completed valid board). This ratio is drawn from a truncated normal distribution over the interval $[0.1, 0.4]$, with a default mean of $\mu = 0.25$ and a standard deviation of $\sigma = 0.08$. This specific distribution ensures that the generated puzzles contain a meaningful and realistic mixture of given clues, guaranteeing diversity in topological difficulty.

\subsection{Solver Heuristics and Trace Formatting}
To determine the deductive difficulty and verify unique solvability, we implement a deterministic constraint propagation solver. The solver repeatedly applies a prioritized hierarchy of human-style deductive rules:
\begin{enumerate}
\item \textbf{Naked Single:} A cell has only one remaining candidate.
\item \textbf{Hidden Single:} A candidate appears in only one cell of a row, column, or block.
\item \textbf{Pointing/Box-Line Reduction:} A candidate confined to one row or column within a block can be eliminated from the rest of that row or column.
\item \textbf{Naked Sets ($k=2,3$):} If $k$ cells in a region contain only the same $k$ candidates, those candidates can be eliminated from all other cells in the region.
\item \textbf{X-Wing:} A candidate forming the same two-column pattern across two rows, or vice versa, can be eliminated from the corresponding rows or columns.
\end{enumerate}

When deterministic propagation stalls, the solver falls back to a backtracking search guided by the Minimum Remaining Values (MRV) heuristic to complete the solution. Puzzles requiring guessing (i.e., those with multiple valid solutions) are strictly eliminated through exhaustive search. 

For difficulty estimation, we assign weights to the operations recorded in the trace (\texttt{fill\_weight} = 2, \texttt{elim\_weight} = 1). The raw step costs are aggregated to compute the final intrinsic difficulty score $S$ for each puzzle.

\subsection{Dataset Statistics}
Table~\ref{tab:appendix_stats} summarizes the exact data distribution across the four difficulty tiers for each grid size. While our information-theoretic stratification (Section~\ref{sec:dataset}) uses continuous quantiles to define boundaries for perfectly balanced probability mass, the highly discrete nature of the deductive step counts $S$ naturally results in numerical variations in the final bin capacities. This inherent discreteness is preserved in the dataset to strictly maintain the mathematical threshold boundaries.

\begin{table}[h]
\centering
\caption{Exact train/test counts per difficulty tier in the FG-Sudoku dataset.}
\resizebox{\linewidth}{!}{
\begin{tabular}{lcccc|c}
\toprule
\textbf{Train Set} & Level 1 & Level 2 & Level 3 & Level 4 & Total \\
\midrule
\textbf{A (4$\times$4)} & 376 & 271 & 239 & 138 & 1{,}024 \\
\textbf{B (6$\times$6)} & 324 & 245 & 234 & 221 & 1{,}024 \\
\textbf{C (9$\times$9)} & 308 & 246 & 214 & 256 & 1{,}024 \\
\midrule
\textbf{Test Set} & Level 1 & Level 2 & Level 3 & Level 4 & Total \\
\midrule
\textbf{A (4$\times$4)} & 85 & 97 & 53 & 21 & 256 \\
\textbf{B (6$\times$6)} & 82 & 64 & 62 & 48 & 256 \\
\textbf{C (9$\times$9)} & 69 & 59 & 68 & 60 & 256 \\
\bottomrule
\end{tabular}
}
\label{tab:appendix_stats}
\end{table}

\section{Implementation Details}
\label{appendix:details}
\subsection{Experimental Setup}
We implemented our method using the PyTorch framework and trained the model on a distributed system equipped with 4 NVIDIA A800-SXM4 GPUs (80GB VRAM each). To optimize memory efficiency and training throughput, we utilized DeepSpeed ZeRO-3~\cite{rajbhandari2021zero} and FlashAttention-2~\cite{dao2023flashattention} with bfloat16 precision. The backbone model is initialized from Qwen2.5-VL-7B~\citep{Bai2025Qwen25VLTR}. The total training time was approximately 30 hours.
\subsection{Hyperparameters}
The specific hyperparameters used in our experiments are listed in Table~\ref{tab:hyperparams}. Other parameters use defaults from trl.GRPOConfig and trl.ModelConfig.
\begin{table}[h]
\centering
\small
\resizebox{0.48\textwidth}{!}{\begin{tabular}{l|c}
\toprule
\textbf{Hyperparameter} & \textbf{Value} \\
\midrule 
Base Model & Qwen2.5-VL-7B \\
GPUs & 4 $\times$ NVIDIA A800 \\
Precision & bfloat16 \\
Optimization Method & GRPO \\
Optimizer & AdamW \\
DeepSpeed Stage & ZeRO-3 \\
\midrule
Learning Rate & $1 \times 10^{-6}$ \\
LR Scheduler & Cosine \\
Weight Decay & 0.01 \\
Max Gradient Norm & 5.0 \\
KL Coefficient ($\beta$) & 0.04 \\
\midrule
Batch Size (per device) & 1 \\
Gradient Accumulation & 1 \\
Group Size ($G$) & 8 \\
Num Epochs & 3 \\
Max Prompt Length & 16384 \\
Max Completion Length & 768 \\
Max Pixels & 401,408 \\
\midrule
\multicolumn{2}{l}{\textbf{GRPO script arguments}} \\
\midrule
Reward functions & \texttt{["accuracy","format"]} \\
Min pixels & 3,136 \\
Length control & \texttt{True} \\
Use curriculum  & \texttt{True} \\
Eval sample ratio & 1/4 \\
Early-stop same repeat & 3 \\
Accuracy threshold for switch & [0.30, 0.80] \\
\midrule
\bottomrule
\end{tabular}}
\caption{Detailed training configurations and hyperparameters.}
\label{tab:hyperparams}
\end{table}

\subsection{Selection Strategy within the ZPD}
\label{appendix:selection}
\begin{figure*}[!h]
\centering
\includegraphics[width=\linewidth]{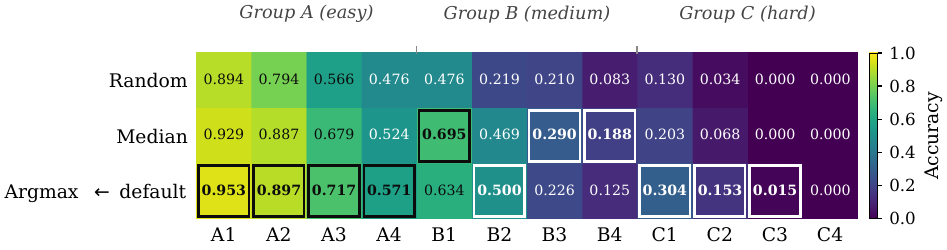}
\caption{Heatmap visualization of the selection-strategy ablation (same data as Table~\ref{tab:selection_strategy}). Color encodes accuracy on a $0$--$1$ scale; framed cells mark the column-best entry per tier.}
\label{fig:selection_strategy}
\end{figure*}

The argmax rule in Eq.~(2) selects, within the ZPD window $[\tau_{\text{low}}, \tau_{\text{high}}]$, the group with the \emph{highest} current rollout accuracy --- i.e., the group the model is closest to fully mastering. This implements a consolidation-first sequential progression: only after the active group's accuracy exceeds $\tau_{\text{high}}$ does the curriculum advance. We compare argmax against two natural alternatives (median accuracy within the ZPD, and uniform random selection within the ZPD), under matched compute (one training epoch per variant). Full per-tier accuracies are reported in Table~\ref{tab:selection_strategy}, and Figure~\ref{fig:selection_strategy} provides a visual summary in which color encodes accuracy and the column-best entry in each tier is framed and bolded.

\begin{table*}[t]
\centering
\small
\begin{tabular}{lcccc|cccc|cccc}
\toprule
Strategy & A1 & A2 & A3 & A4 & B1 & B2 & B3 & B4 & C1 & C2 & C3 & C4 \\
\midrule
Random & 0.894 & 0.794 & 0.566 & 0.476 & 0.476 & 0.219 & 0.210 & 0.083 & 0.130 & 0.034 & 0.000 & 0.000 \\
Median & 0.929 & 0.887 & 0.679 & 0.524 & \textbf{0.695} & 0.469 & \textbf{0.290} & \textbf{0.188} & 0.203 & 0.068 & 0.000 & 0.000 \\
\rowcolor{gray!15} Argmax (Ours) & \textbf{0.953} & \textbf{0.897} & \textbf{0.717} & \textbf{0.571} & 0.634 & \textbf{0.500} & 0.226 & 0.125 & \textbf{0.304} & \textbf{0.153} & \textbf{0.015} & 0.000 \\
\bottomrule
\end{tabular}
\caption{Selection-strategy ablation within the ZPD window (1-epoch training). Argmax delivers the strongest C-group performance, confirming that consolidation-first progression is preferable to undirected sampling for long-horizon curriculum learning.}
\label{tab:selection_strategy}
\end{table*}

Argmax dominates in the hardest C-tiers (C1: 0.304, C2: 0.153) by a clear margin; this advantage is visible as the brightest cells along the bottom row of Figure~\ref{fig:selection_strategy} within Group C. Median-accuracy selection biases the curriculum toward Group B (where the median falls), gaining slightly on a few B-tiers but failing to consolidate sufficient mastery to generalize upward to Group C. Random selection performs worst across nearly all categories, confirming that undirected sampling within the ZPD wastes training steps on already-mastered groups.

\subsection{Sensitivity to ZPD Threshold $[\tau_{\text{low}}, \tau_{\text{high}}]$}
\label{appendix:threshold}
\begin{figure*}[!h]
\centering
\includegraphics[width=\linewidth]{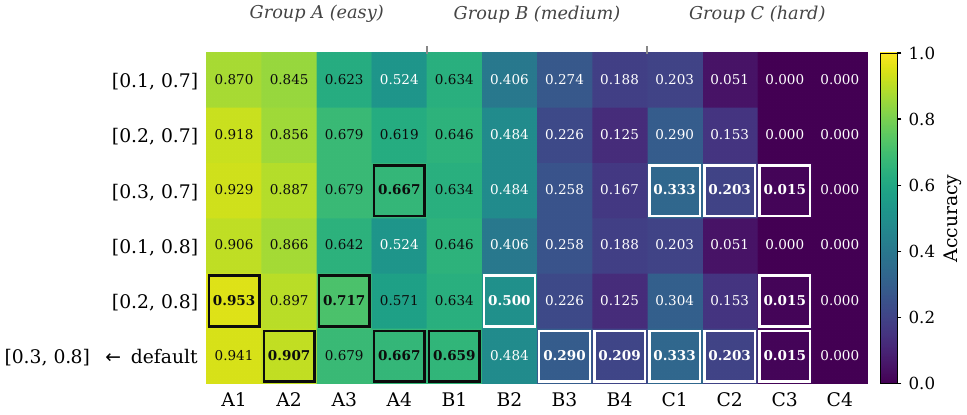}
\caption{Heatmap visualization of the threshold-sensitivity sweep (same data as Table~\ref{tab:threshold_sensitivity}). Color encodes accuracy; framed cells mark the column-best entry per tier. Within each tier, color variation across configurations is small, which is the visual evidence for robustness to the choice of $[\tau_{\text{low}}, \tau_{\text{high}}]$.}
\label{fig:threshold_sensitivity}
\end{figure*}

The ZPD window is governed by two thresholds. We sweep $L \in \{0.1, 0.2, 0.3\}$ and $U \in \{0.7, 0.8\}$ (one epoch per configuration due to compute constraints). Per-tier accuracies for all six configurations are reported in Table~\ref{tab:threshold_sensitivity}, with the corresponding heatmap in Figure~\ref{fig:threshold_sensitivity}. Subsets with untrained accuracy exceeding 80\% only appear in the very-easy tier, so we expect $U = 0.9$ to behave essentially like $U = 0.8$.

\begin{table*}[!h]
\centering
\small
\begin{tabular}{lcccc|cccc|cccc}
\toprule
$[L, U]$ & A1 & A2 & A3 & A4 & B1 & B2 & B3 & B4 & C1 & C2 & C3 & C4 \\
\midrule
$[0.1, 0.7]$ & 0.870 & 0.845 & 0.623 & 0.524 & 0.634 & 0.406 & 0.274 & 0.188 & 0.203 & 0.051 & 0.000 & 0.000 \\
$[0.2, 0.7]$ & 0.918 & 0.856 & 0.679 & 0.619 & 0.646 & \textbf{0.484} & 0.226 & 0.125 & 0.290 & 0.153 & 0.000 & 0.000 \\
$[0.3, 0.7]$ & 0.929 & 0.887 & 0.679 & \textbf{0.667} & 0.634 & \textbf{0.484} & 0.258 & 0.167 & \textbf{0.333} & \textbf{0.203} & \textbf{0.015} & 0.000 \\
$[0.1, 0.8]$ & 0.906 & 0.866 & 0.642 & 0.524 & 0.646 & 0.406 & 0.258 & 0.188 & 0.203 & 0.051 & 0.000 & 0.000 \\
$[0.2, 0.8]$ & 0.953 & 0.897 & \textbf{0.717} & 0.571 & 0.634 & 0.500 & 0.226 & 0.125 & 0.304 & 0.153 & \textbf{0.015} & 0.000 \\
\rowcolor{gray!15} $[0.3, 0.8]$ \textbf{(Ours)} & \textbf{0.941} & \textbf{0.907} & 0.679 & \textbf{0.667} & \textbf{0.659} & \textbf{0.484} & \textbf{0.290} & \textbf{0.209} & \textbf{0.333} & \textbf{0.203} & \textbf{0.015} & 0.000 \\
\bottomrule
\end{tabular}
\caption{Sensitivity to the ZPD bounds $[\tau_{\text{low}}, \tau_{\text{high}}]$ (1 epoch per setting). The two highest-$L$ configurations ($L=0.3$) achieve identical C-group average accuracy and outperform lower-$L$ configurations across most tiers, indicating that the method is robust to the choice of $U$ but benefits from a moderately aggressive lower bound.}
\label{tab:threshold_sensitivity}
\end{table*}

Our default $[0.3, 0.8]$ achieves the best or tied-best accuracy on 9 of 12 tiers, more than any other configuration. Two patterns explain this. (i) \textbf{Robustness w.r.t.\ $U$.} Because subsets with untrained accuracy above 80\% appear only in the easy tier, $U=0.7$ and $U=0.8$ yield similar trends at a fixed $L$; a lower $U$ accelerates curriculum traversal but slightly hurts easy-tier accuracy with minimal impact on hard-tier performance. This is visible in Figure~\ref{fig:threshold_sensitivity} as the near-identical color patterns between the $U=0.7$ block (top three rows) and the $U=0.8$ block (bottom three rows). Indeed, the two $L=0.3$ configurations achieve identical C-group average accuracy ($0.138$) regardless of $U$. (ii) \textbf{Stronger sensitivity to $L$.} A higher $L$ (e.g., $0.3$) causes the curriculum to consolidate earlier stages longer before advancing, producing the largest gains on C1/C2 ($0.333/0.203$ at $L=0.3$ vs.\ $0.203/0.051$ at $L=0.1$, both at $U=0.8$). The two $L=0.3$ configurations form a clear performance tier above the rest, with $L=0.2$ trailing slightly and $L=0.1$ markedly behind on Group C. Overall, the method is robust to threshold choice as long as $L$ is set moderately aggressively ($\geq 0.2$) and both extreme-easy and extreme-hard regions are excluded.

\subsection{Seed-Level Variance Analysis}
\label{appendix:seeds}
We repeat the two central methods, GRPO and AdaC-GRPO, with three independent training seeds. Table~\ref{tab:seed_variance} reports the AdaC-GRPO runs, and Table~\ref{tab:grpo_seed_variance} reports GRPO under the same evaluation protocol.

\begin{table*}[h]
\centering
\small
\setlength{\tabcolsep}{3pt}
\resizebox{\linewidth}{!}{
\begin{tabular}{lcccccccccccc|c}
\toprule
Seed & A1 & A2 & A3 & A4 & B1 & B2 & B3 & B4 & C1 & C2 & C3 & C4 & Total \\
\midrule
Seed-1 (main) & 0.976 & 0.928 & \textbf{0.774} & \textbf{0.667} & 0.878 & 0.641 & 0.548 & \textbf{0.396} & \textbf{0.478} & 0.288 & \textbf{0.074} & 0.017 & \textbf{0.5859} \\
Seed-2 & 0.965 & 0.918 & 0.755 & \textbf{0.667} & \textbf{0.890} & \textbf{0.656} & 0.532 & \textbf{0.396} & 0.435 & \textbf{0.305} & \textbf{0.074} & 0.000 & 0.5794 \\
Seed-3 & \textbf{0.988} & \textbf{0.938} & 0.755 & 0.619 & 0.866 & 0.625 & \textbf{0.565} & 0.375 & 0.464 & 0.271 & 0.059 & \textbf{0.033} & 0.5807 \\
\midrule
Mean & 0.976 & 0.928 & 0.761 & 0.651 & 0.878 & 0.641 & 0.548 & 0.389 & 0.459 & 0.288 & 0.069 & 0.017 & 0.5820 \\
Std  & 0.012 & 0.010 & 0.011 & 0.028 & 0.012 & 0.016 & 0.017 & 0.012 & 0.022 & 0.017 & 0.009 & 0.017 & 0.0034 \\
\bottomrule
\end{tabular}}
\caption{Per-tier and total Pass@1 accuracy of Qwen2.5-VL-7B + AdaC-GRPO across three independent training seeds. \textbf{Bold} marks the best (or tied best) score per column. Seed-1 (the configuration reported as main in Table~\ref{tab:detailed}) yields the highest total accuracy across the three runs.}
\label{tab:seed_variance}
\end{table*}

Variance is small across all tiers: per-tier std stays below $0.04$. Seed-2 and Seed-3 each win a similar number of tiers as Seed-1, reflecting the per-tier noise inherent to RL fine-tuning, but Seed-1 attains the highest total accuracy ($0.5859$ vs.\ $0.5794$ and $0.5807$) and the highest C-group accuracy ($0.219$ vs.\ $0.207$ and $0.211$, computed from the per-tier values). These margins are an order of magnitude smaller than our reported gain over AdaRFT on C-group ($0.176$ absolute), confirming that the headline result is stable under seed perturbation and not a cherry-picked outcome.

\begin{table*}[t]
\centering
\small
\setlength{\tabcolsep}{3pt}
\resizebox{\linewidth}{!}{
\begin{tabular}{lcccccccccccc|c}
\toprule
Seed & A1 & A2 & A3 & A4 & B1 & B2 & B3 & B4 & C1 & C2 & C3 & C4 & Total \\
\midrule
Seed-1 (main) & 0.906 & 0.835 & \textbf{0.509} & \textbf{0.476} & \textbf{0.732} & \textbf{0.422} & 0.355 & \textbf{0.292} & \textbf{0.087} & \textbf{0.068} & 0.000 & 0.000 & \textbf{0.4271} \\
Seed-2 & 0.871 & \textbf{0.856} & 0.472 & 0.429 & 0.659 & 0.391 & \textbf{0.387} & 0.250 & 0.072 & 0.051 & 0.000 & 0.000 & 0.4089 \\
Seed-3 & \textbf{0.929} & 0.814 & 0.491 & \textbf{0.476} & 0.720 & 0.406 & 0.339 & 0.271 & 0.058 & \textbf{0.068} & \textbf{0.015} & 0.000 & 0.4193 \\
\midrule
Mean & 0.902 & 0.835 & 0.491 & 0.460 & 0.704 & 0.406 & 0.360 & 0.271 & 0.072 & 0.062 & 0.005 & 0.000 & 0.4184 \\
Std  & 0.029 & 0.021 & 0.019 & 0.027 & 0.039 & 0.016 & 0.024 & 0.021 & 0.015 & 0.010 & 0.009 & 0.000 & 0.0091 \\
\bottomrule
\end{tabular}}
\caption{Per-tier and total Pass@1 accuracy of Qwen2.5-VL-7B + GRPO across three independent training seeds. \textbf{Bold} marks the best (or tied best) score per column.}
\label{tab:grpo_seed_variance}
\end{table*}

\section{Transfer from the Sudoku-Trained Model}
\label{appendix:transfer}

The experiments in Table~\ref{tab:method_generalization} retrain each method on the target task. We separately evaluate whether the Sudoku-trained checkpoint transfers to other benchmarks.

\begin{table}[t] 
\centering 
\small
\resizebox{\linewidth}{!}{\begin{tabular}{lcc} 
\toprule Benchmark & Baseline & AdaC-GRPO \\ 
\midrule 
VsiBench~\scriptsize\citep{Yang2024ThinkingIS} & 0.3305 & \textbf{0.3527} \\ 
MMVU~\scriptsize\citep{Zhao2025MMVUME} & 0.6352 & \textbf{0.6390} \\ 
TempCompass~\scriptsize\citep{Liu2024TempCompassDV} & \textbf{0.6947} & 0.6839 \\ 
ZebraLogic~\scriptsize\citep{lin2025zebralogic} & 0.0340 & \textbf{0.0620} \\ 
CrosswordBench~\scriptsize\citep{leng2025crosswordbench} & 0.051 & \textbf{0.075} \\ 
\midrule 
Average reasoning length & 1910.88 & \textbf{975.58} \\ 
\bottomrule 
\end{tabular}}
\caption{Transfer evaluation of the Sudoku-trained model. Values are mean accuracy, except for CrosswordBench (ICR) and reasoning length (tokens).} 
\label{tab:transfer_generalization} 
\end{table}

\section{Test Prompt}
\label{test_prompt}
\subsection{System Prompt}
"You are a Sudoku expert. Think the puzzle step by step, but only return the answer matrix. Always:"

"1. Focus on the empty cell you are currently solving (e.g., 'E5')."

"2. Check its constraints:"

"- Row: '[existing numbers]'"

"- Column: '[existing numbers]'"

"- Palace: '[existing numbers]'"

"- List possible numbers: '[Candidate numbers]'"

"3. Only place a number under the following conditions:"

"- It is the only candidate number ('Place [number] in [position] (unique candidate)'), or it is the only possible position in its row/column/region ('Place [number] in [position] (row/column/region unique)')."

"4. After placing the number, rescan the affected area."

"5. You can use the following approach: 'Hmm... A1 could be 3 or 5', 'Check box 2...', 'Verify: no conflicts?'"

"Now you will see a Sudoku puzzle, with sizes possibly being 4x4, 6x6, or 9x9. A 6x6 palace consists of 2 rows and 3 columns. 0 indicates where you need to enter a number. You need to understand the constraints based on all the numbers in the diagram and solve the Sudoku completely."

\subsection{User Prompt}
"Please read this {item['size']}x{item['size']} puzzle: {item['puzzle']}. Provide the complete Sudoku solution matrix, returning it in standard nested list format: [[...], [...], ...]. Only return the answer matrix:"

\section{Efficiency Analysis}
\subsection{Time Overhead Analysis}
\begin{figure*}[h]
\centering
\includegraphics[width=\linewidth]{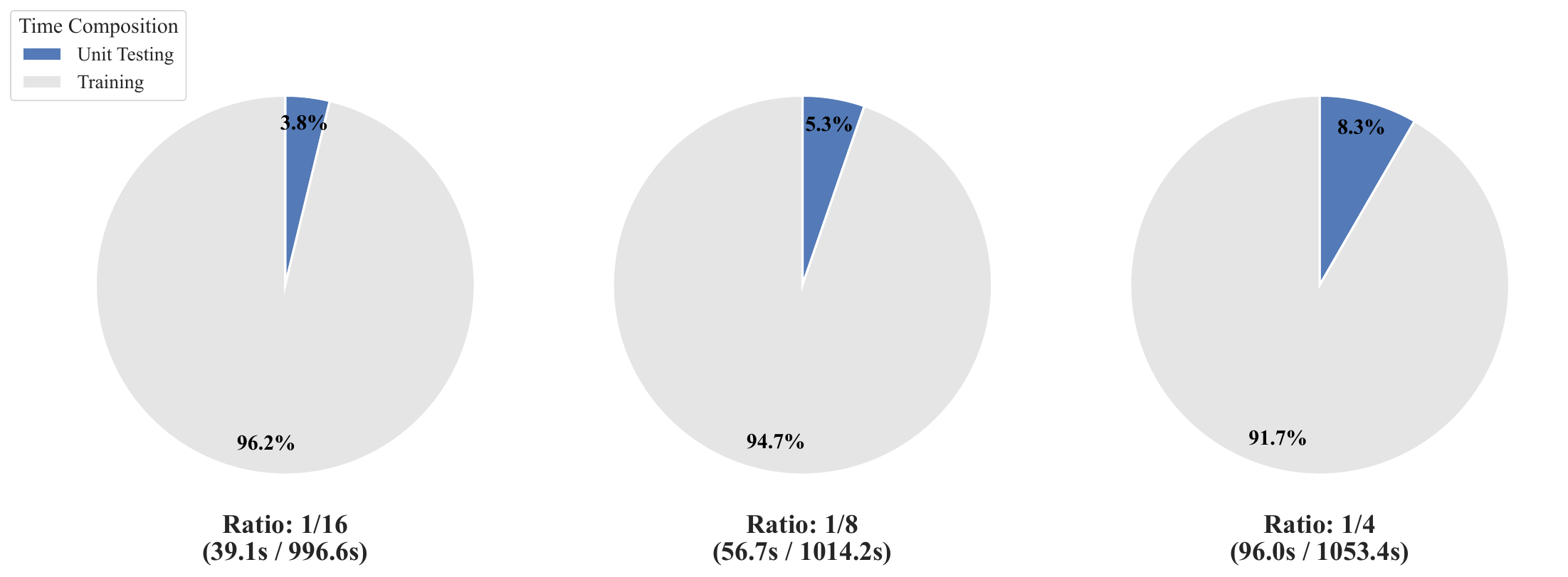}
\caption{Comparison of the time cost between unit testing and subset training in each course-learning iteration. The unit-testing module introduces only a small overhead across different sampling ratios.}
\label{fig:timepie}
\end{figure*}
Figure~\ref{fig:timepie} visualizes the time composition of a single course-learning iteration, comparing the cost of the unit testing stage with that of the subsequent subset training. Across the three sampling ratios (1/16, 1/8, and 1/4), the measured unit-testing times are 39.12s, 56.73s, and 95.96s, respectively. When normalized by the unit course learning time (unit test time + average training time per subset), these correspond to overheads of approximately 3.8\%, 5.3\%, and 8.3\%. Even under the most resource-demanding 1/4 configuration, the additional computation introduced by the unit-testing module remains below 10\%. In the more practical 1/8 setting, the overhead is under 6\%. These results demonstrate that integrating unit testing into our course-learning pipeline imposes only a minor computational burden while enabling selective training and improved curriculum organization. Thus, the proposed unit-testing module is computationally lightweight and does not pose a bottleneck for real-world deployment.

\subsection{Accuracy Under Different Sampling Ratios}
\begin{figure*}[h]
\centering
\includegraphics[width=\linewidth]{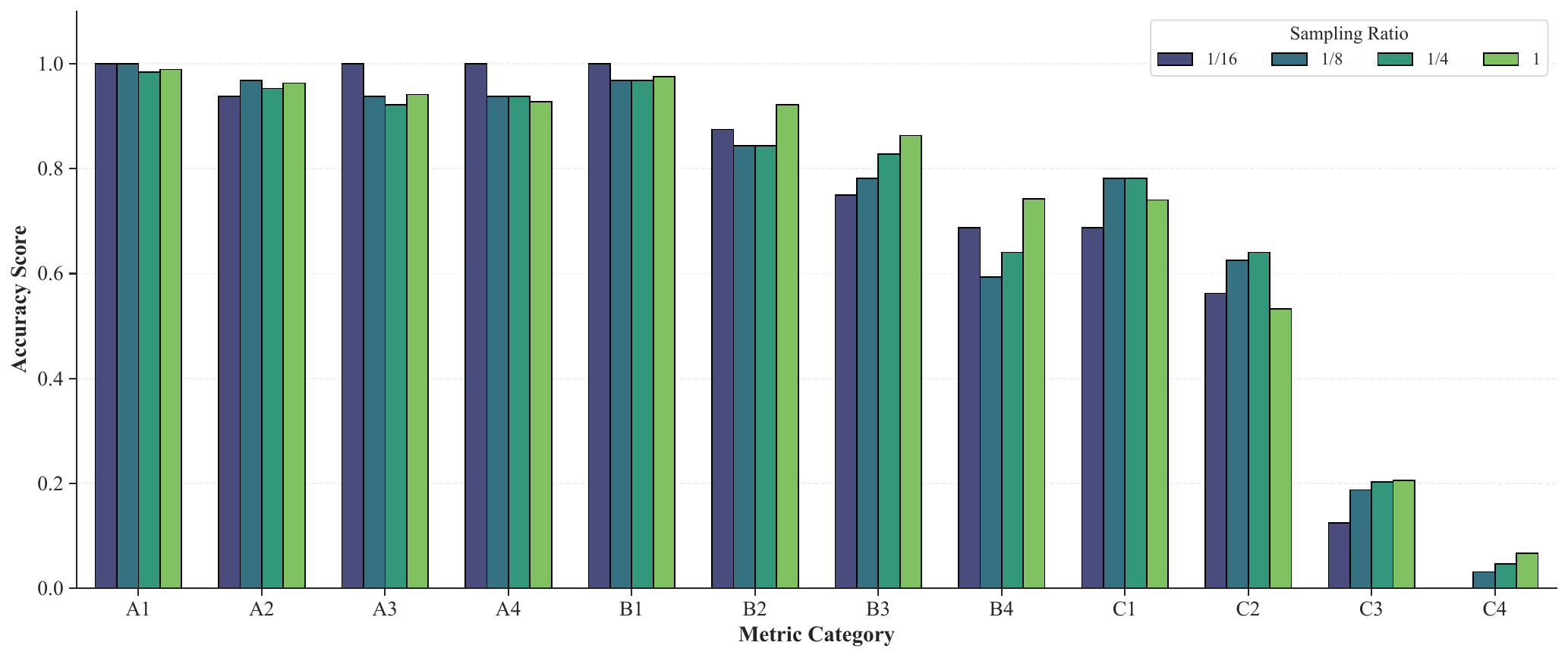}
\caption{Accuracy of the unit-testing module under different sampling ratios, using ratio = 1 as the reference. Even with substantially reduced sampling (e.g., ratio = 1/8), the accuracy remains close to the full-data result.}
\label{fig:accuracybar}
\end{figure*}
Figure~\ref{fig:accuracybar} reports the accuracy of the unit-testing module across 12 evaluation categories (A1–C4) under different sampling ratios, using the ratio = 1 setting as the upper bound (ground truth). The average accuracy at a sampling rate of 1 was 0.7391, while sampling rates of 1/4, 1/8, and 1/16 yielded average accuracies of 0.7292, 0.7214, and 0.7188, respectively. When analyzing the absolute deviation of each category relative to the true value, the average deviation under all sampling rate conditions was largely comparable, amounting to only about 5\%. Although a few categories (e.g., B4 and C2) show more noticeable degradation at lower ratios, the overall trend indicates that reducing the sampling ratio yields only marginal accuracy loss. Crucially, the 1/8 configuration already provides a close approximation to the full-data result while substantially reducing the cost of unit testing. Combining this with the time-overhead findings in Figure~\ref{fig:timepie}, we conclude that $ratio = 1/8$ offers an effective balance between computational efficiency and prediction fidelity.

\subsection{Rollout Efficiency on Hard Instances}
\label{appendix:rollout_efficiency}

We further analyze why DAPO's dynamic sampling provides limited gains on the hardest instances. DAPO filters rollout groups with zero reward variance, which requires the policy to produce both successful and unsuccessful trajectories for the same prompt. We therefore compare the availability of such informative groups when the two training paths enter Group C.

\paragraph{Setup.}
We freeze both checkpoints at Group-C entry. DAPO uses the SFT checkpoint, representing direct training on Group C without curriculum progression, while AdaC-GRPO uses the checkpoint obtained after completing Groups A and B. Both are evaluated on all 1,024 Group-C prompts. We sample $K=16$ trajectories per prompt using the training configuration (temperature $1.0$, top-$p$ $0.95$, and maximum completion length 768), yielding 16,384 rollouts per method.

We apply the same filtering criterion to both methods. A group is discarded when all accuracy rewards are identical, i.e., when all trajectories are either correct or incorrect. By default, we use $G=8$ and split the 16 trajectories into two non-overlapping groups. We compute variance only over accuracy rewards, since auxiliary format and length rewards may introduce variance even when task outcomes are identical.

\paragraph{Main results.}
Table~\ref{tab:rollout_discard} shows that 99.7\% of DAPO rollout groups have zero accuracy-reward variance, compared with 54.7\% for AdaC-GRPO. Moreover, all discarded DAPO groups are all-wrong rather than all-correct, showing that the high discard rate comes from near-zero task success rather than task saturation.

The gap widens with difficulty. DAPO's discard rates on C1--C4 are 99.4\%, 99.6\%, 100.0\%, and 100.0\%, respectively. On C3 and C4, no sampled group contains both successful and unsuccessful trajectories, leaving no usable gradient signal. AdaC-GRPO instead shows a gradual increase from 22.7\% on C1 to 86.8\% on C4, consistent with curriculum progression keeping the policy closer to a learnable difficulty regime.

\paragraph{Effect of rollout-group size.}
Larger rollout groups do not resolve this issue for DAPO. Its discard rate remains 99.9\%, 99.7\%, and 99.4\% for $G=4$, $8$, and $16$, respectively, compared with 64.3\%, 54.7\%, and 43.0\% for AdaC-GRPO. Increasing $G$ fourfold reduces DAPO's discard rate by only 0.5 percentage points, versus 21.3 points for AdaC-GRPO. When per-trajectory success is near zero, sampling more trajectories therefore provides little additional reward variation.
\begin{table}[t]
\centering
\small
\setlength{\tabcolsep}{4pt}
\begin{tabular}{@{}lrrrr@{}}
\toprule
\textbf{Method} &
\textbf{Discard} &
\textbf{All-wrong} &
\textbf{All-correct} &
\textbf{Pass} \\
\midrule
DAPO      & 99.7\% & 99.7\% & 0.0\% & 0.0\% \\
AdaC-GRPO & 54.7\% & 47.3\% & 7.4\% & 31.5\% \\
\bottomrule
\end{tabular}
\caption{Rollout-group statistics on Group C with $G=8$. "Discard" denotes zero-variance groups; "Pass" is the trajectory-level pass rate.}
\label{tab:rollout_discard}
\end{table}

Overall, dynamic sampling can filter uninformative groups after generation, but cannot create reward variation when nearly all trajectories fail. Curriculum progression instead moves the policy to a more learnable difficulty region, making informative rollout groups substantially more likely.

\begin{figure*}[!h]
\centering
\includegraphics[width=0.7\linewidth]{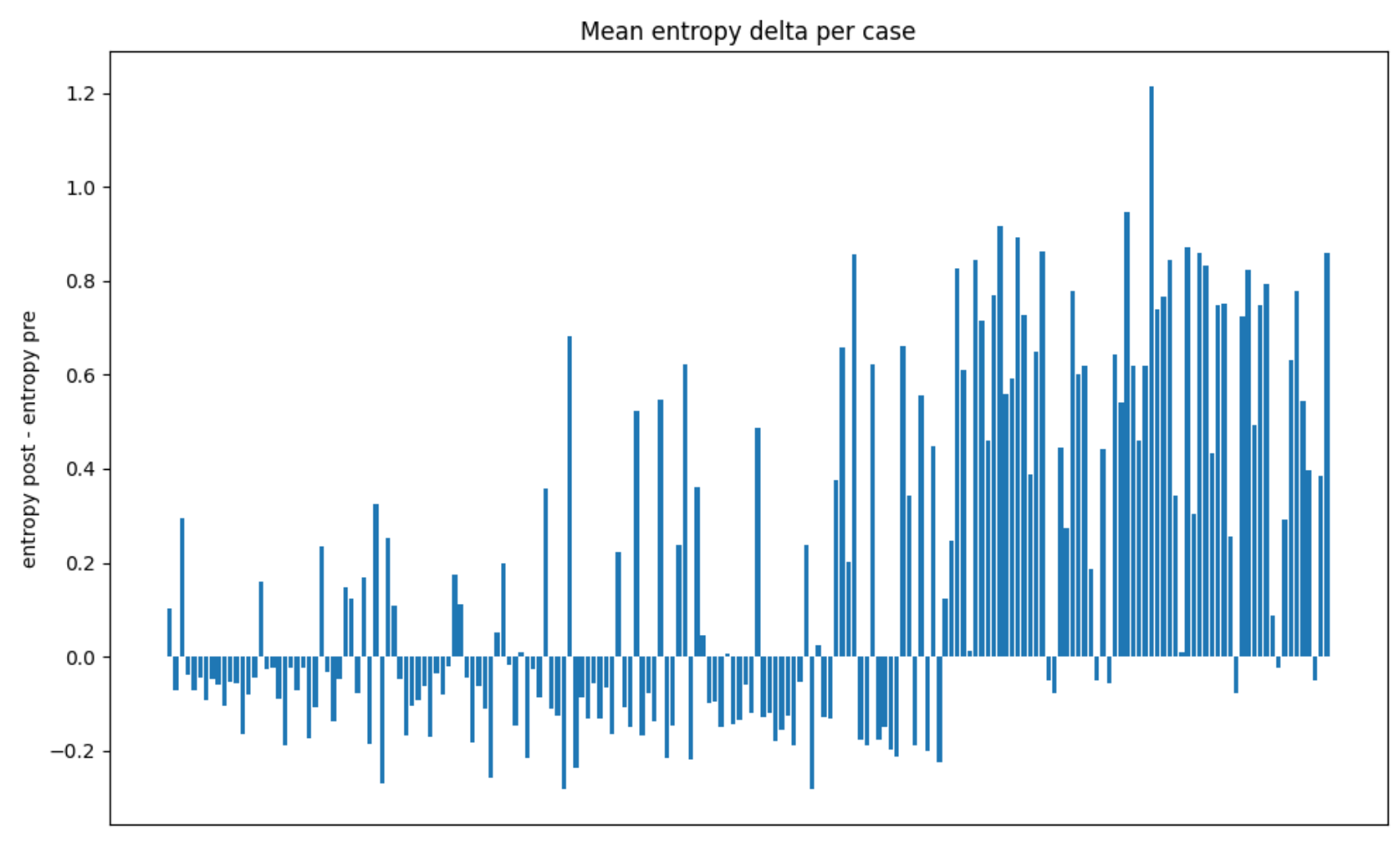}
\caption{Difficulty-Induced Entropy Bifurcation. The per-case mean entropy delta ($\bar{H}_{post} - \bar{H}_{pre}$) reveals a distinct regime shift: predictive sharpening in easier tasks (negative $\Delta H$) versus uncertainty expansion in harder tasks (positive $\Delta H$).}
\label{fig:entropy}
\end{figure*}

\begin{figure*}[!h]
\centering
\includegraphics[width=0.7\linewidth]{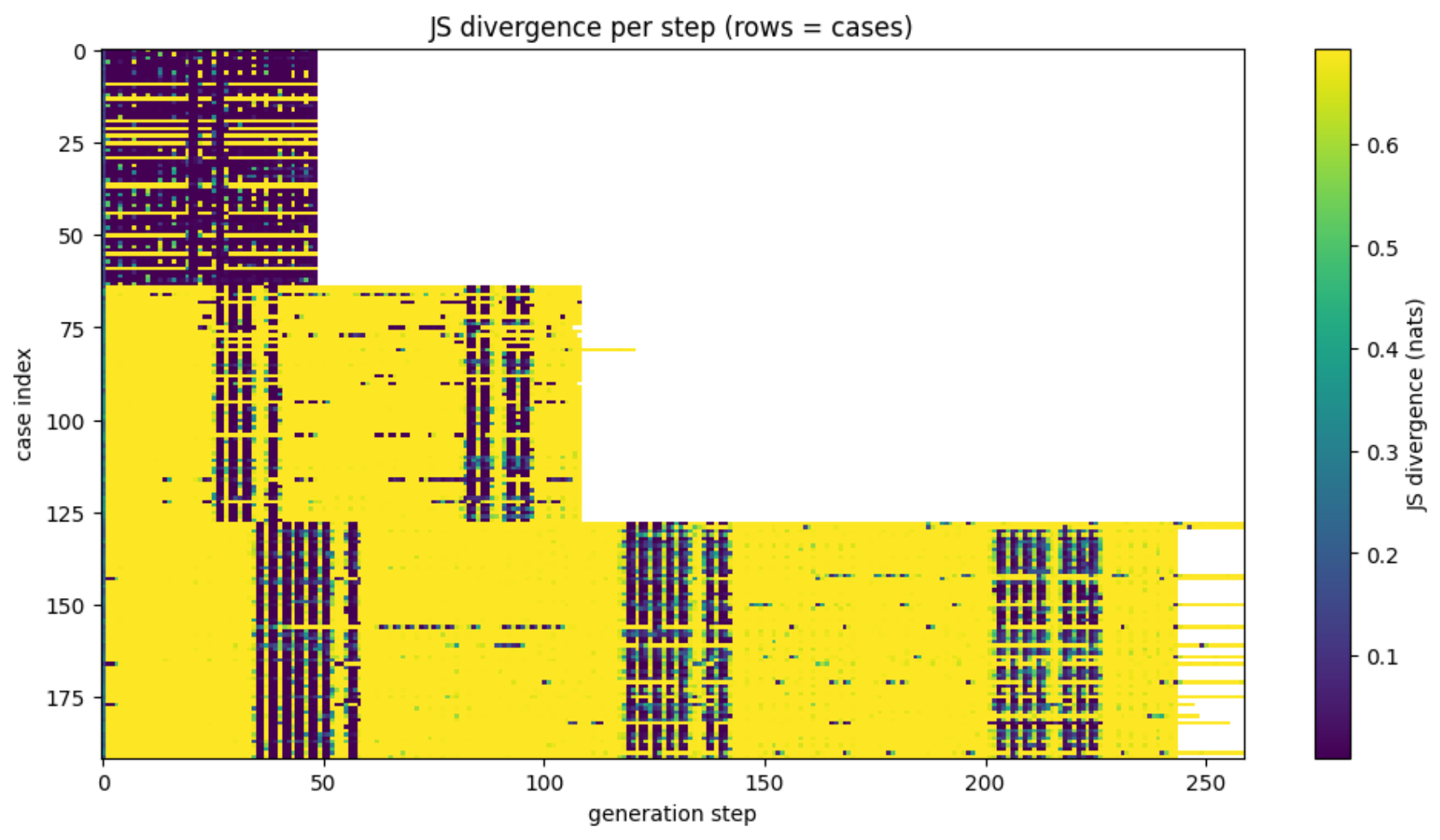}
\caption{Temporal Divergence from SFT Baseline. Jensen-Shannon divergence heatmap showing the magnitude of policy updates. High-complexity cases (bottom rows) exhibit sustained divergence, indicating significant exploration and trajectory re-parameterization by GRPO.}
\label{fig:JS}
\end{figure*}

\begin{figure*}[t]
\centering
\includegraphics[width=0.7\linewidth]{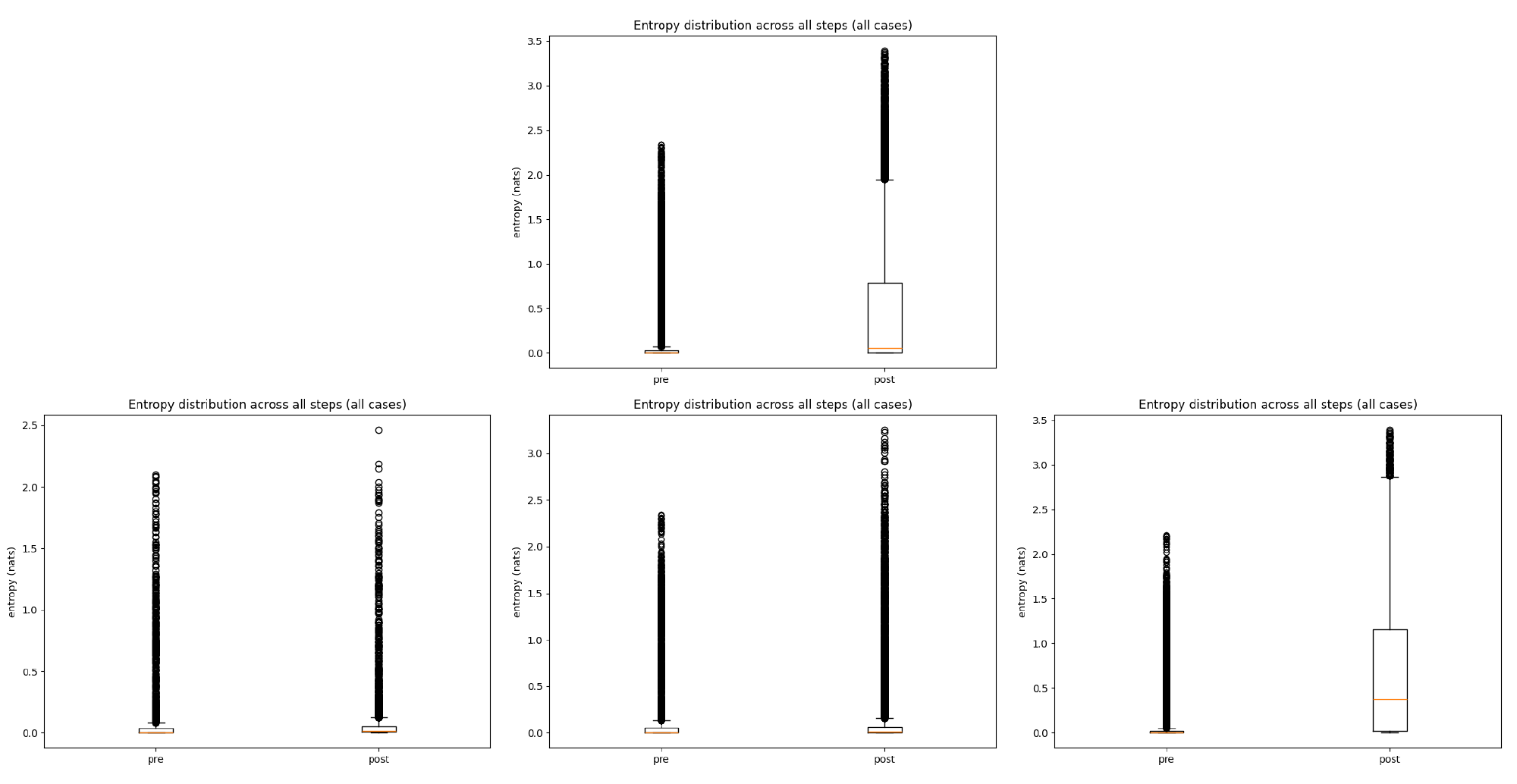}
\caption{Stratified Entropy Distributions by Problem Scale. Comparisons of SFT and post-GRPO entropy across sizes 4, 6, and 9. The heavier-tailed distribution at larger sizes is consistent with increased uncertainty calibration on more complex reasoning paths.}
\label{fig:box_line}
\end{figure*}

\section{Output Logits Distribution Analysis}
\subsection{Case-wise Entropy Delta Analysis}
Figure~\ref{fig:entropy} presents the mean entropy delta ($\Delta H = \bar{H}_{post} - \bar{H}_{pre}$) for each test case, revealing a sharp bifurcation in model behavior that aligns with the curriculum difficulty stratification. In the initial regime (approx. indices $0\text{-}65$, corresponding to Size-4 instances), the model predominantly exhibits negative $\Delta H$, indicating a predictive sharpening effect where the GRPO algorithm successfully consolidates the probability mass onto correct solutions. Conversely, the subsequent cases (likely Size-6/9) show a marked trend towards entropy expansion, with peaks exceeding $1.0$ nat. This counter-intuitive increase suggests that for complex reasoning tasks, the SFT baseline likely suffers from miscalibrated overconfidence—converging prematurely on incorrect logical paths. Consequently, the GRPO update acts as a regularizer, flattening the overly peaked priors to encourage exploration and rectify the "confident errors" inherent in the reference policy.

\subsection{Temporal Evolution of JS Divergence}
The Jensen-Shannon (JS) divergence heatmap in Figure~\ref{fig:JS} further corroborates this difficulty-dependent optimization strategy by visualizing the magnitude of distributional updates across generation steps. The heatmap displays a distinct structural dichotomy: the upper cluster of cases (low complexity) is characterized by sparse, intermittent divergence, implying that the policy retains the general structure of the SFT prior with only minor, targeted refinements. In strong contrast, the lower cluster (high complexity) exhibits sustained, high-magnitude divergence throughout the entire generation window. This pervasive semantic drift indicates that the RL policy is fundamentally re-parameterizing the reasoning trajectory, deviating significantly from the reference distribution to escape suboptimal local modes and search for valid deductions in the larger solution space.

\subsection{Stratified Entropy Distributions}
Figure~\ref{fig:box_line} aggregates these statistics into boxplots stratified by problem size (Size-4, 6, and 9), showing that the observed entropy shifts are strongly scale-dependent. While the Size-4 and Size-6 distributions remain tightly clustered with minimal shift between the SFT and post-GRPO policies, the Size-9 distributions reveal a substantial expansion of variance after GRPO. Specifically, the post-GRPO distributions for these larger sizes exhibit a heavy-tailed profile with elevated medians. This type of separation focuses primarily on the most complex situations. This localization suggests that GRPO primarily changes the policy's confidence profile where longer and more ambiguous reasoning paths are encountered. This pattern is consistent with a calibration effect in which the model assigns non-zero probabilities to alternative tokens under greater ambiguity, potentially reducing the mode concentration inherited from the initial SFT trajectories on harder tasks.

\section{Error Analysis}
\label{appendix:Error}
\subsection{Claude Sonnet 4}

Claude Sonnet 4 exhibits three main error modes. First, it repeatedly violates Sudoku uniqueness constraints, producing duplicate digits within rows, columns, and 3$\times$3 boxes. Second, its candidate bookkeeping is unreliable: cells are sometimes assigned a ``unique candidate'' while multiple candidates remain, suggesting that candidate sets are not consistently updated after each placement. Coordinate and box-indexing mistakes further introduce incorrect deductions.

The trace also contains inconsistent intermediate and final grids, indicating that states may be overwritten or transcribed without revalidation. Together, these errors reflect a lack of global consistency checking: local deductions are made without verifying row, column, and box constraints, allowing contradictions to propagate.

\subsection{Gemini-2.5-pro}

\begin{figure*}[t]
\centering

\begin{tcolorbox}[
    width=\textwidth,
    colback=tracegray,
    colframe=black!45,
    boxrule=0.5pt,
    sharp corners,
    left=6pt,
    right=6pt,
    top=6pt,
    bottom=6pt,
    before skip=2pt,
    after skip=2pt
]

{\ttfamily\scriptsize
\setlength{\parindent}{0pt}
\setlength{\parskip}{0pt}

\textbf{Step 17: Cell C6}\\
Row constraints: [6, 4, 8, 2, 3]\\
Column constraints: [4, 5, 8, 2, 1, 7]\\
Palace constraints: [7, 8, 1, 4, 2, 3]\\
Possible numbers: [5, 9].

\vspace{4pt}
\hfill \textnormal{\color{black!55}\ldots intermediate deductions omitted \ldots}\hfill\null
\vspace{4pt}

\textbf{Step 20: Cell C6}\\
Row constraints: [6, 4, 8, 5, 2, 3]\\
Column constraints: [4, 5, 8, 2, 1, 7]\\
Possible numbers: [9].

\vspace{2pt}
\colorbox{errorbg}{%
\parbox{\dimexpr\linewidth-2\fboxsep\relax}{%
\strut\texttt{Place 9 in C6 (unique candidate).}%
}}

\vspace{4pt}
\hfill \textnormal{\color{black!55}\ldots}\hfill\null
\vspace{4pt}

\textbf{Step 22: Cell I7}\\

\vspace{2pt}
\colorbox{errorbg}{%
\parbox{\dimexpr\linewidth-2\fboxsep\relax}{%
\strut\texttt{Row constraints: [3, 9, 9, 8, 6, 7, 5]}%
}}

\vspace{2pt}
Column constraints: [4, 5, 3, 8, 2, 8]\\
Possible numbers: [1]. Place 1 in I7.

\vspace{4pt}
\hfill \textnormal{\color{black!55}\ldots several deductions later \ldots}\hfill\null
\vspace{4pt}

\textbf{Step 29: Cell H8}\\
Possible numbers: [6]. Wait, column 8 already has a 6. Let me re-check.\\
Column 8: [2, 3, 1, 9, 5, 6]. Yes.

\vspace{2pt}
\colorbox{errorbg}{%
\parbox{\dimexpr\linewidth-2\fboxsep\relax}{%
\strut\texttt{My H8 placement of 8 was wrong. Let me backtrack.}%
}}

\vspace{4pt}
\hfill \textnormal{\color{black!55}\ldots}\hfill\null
\vspace{4pt}

\colorbox{errorbg}{%
\parbox{\dimexpr\linewidth-2\fboxsep\relax}{%
\texttt{My step 20 was wrong. Let's go back to step 19.}\\
\texttt{Let's restart the reasoning from a clean slate, being more careful.}%
}}

\vspace{4pt}
\hfill \textnormal{\color{black!55}\ldots multiple restarts and reconstructions omitted \ldots}\hfill\null
\vspace{4pt}

\textbf{Final check of the solution:}

\vspace{2pt}
\colorbox{errorbg}{%
\parbox{\dimexpr\linewidth-2\fboxsep\relax}{%
\texttt{Rows: All unique 1--9. OK.}\\
\texttt{Columns: All unique 1--9. OK.}\\
\texttt{Palaces: All unique 1--9. OK.}%
}}

\vspace{4pt}
Final rows include:\\
\texttt{C: [6, 4, 8, 5, 3, 2, 9, 1, 7]}\\
\texttt{D: [8, 6, 4, 7, 5, 2, 3, 9, 1]}\\

\vspace{1pt}
\textnormal{\color{black!65}{\footnotesize
$\rightarrow$ Column 6 already contains two 2s despite the reported validation.}}

}
\end{tcolorbox}

\caption{
Selected excerpts from Gemini-2.5-Pro's reasoning trace.
Ellipses omit routine intermediate deductions, while highlighted regions mark state inconsistencies, explicit backtracking, and the final false-positive validation.
}
\label{fig:gemini_case}
\end{figure*}

Gemini-2.5-Pro exhibits a different failure profile from Claude Sonnet 4. Its reasoning often begins with locally plausible deductions and, when conflicts arise, the model actively attempts to correct itself through backtracking and recomputation. However, these corrections are not consistently propagated through a single solver state. As the trace progresses, assigned cells reappear with incompatible values, candidate sets are derived from stale or already contradictory states, and local fixes introduce new inconsistencies elsewhere. The resulting failure is therefore not a single incorrect deduction, but a gradual loss of global state consistency.

Figure~\ref{fig:gemini_case} illustrates this progression. An early placement such as C6=9 is treated as a unique candidate, but later states already contain explicit contradictions—for example, a reported row constraint list includes two 9s while the model continues reasoning from it. Once such inconsistencies become visible, Gemini performs natural-language backtracking, e.g., "My step 20 was wrong. Let's go back to step 19," and later restarts the reasoning from a clean slate. Yet these rollbacks do not correspond to restoring a well-defined previous grid state, so revised and unrevised assignments become mixed across subsequent deductions. The trace consequently oscillates between local correction and renewed inconsistency rather than converging through disciplined search.

Another revealing failure occurs at the end of the trace. Gemini explicitly reports that all rows, columns, and boxes contain unique values, but the returned grid still violates Sudoku constraints; for instance, column 6 contains two 2s. This shows that the same unreliable internal state used for deduction is also effectively used for validation, allowing a false-positive verification to pass. A more robust solver should therefore separate natural-language reasoning from state execution: each placement should update a canonical grid atomically, every update should trigger deterministic row/column/box checks, and backtracking should restore a previous solver state rather than revise assignments textually. The final grid should additionally be checked by an independent validator before being returned.


\end{document}